\documentclass{article}

\usepackage{microtype}
\usepackage{graphicx}
\usepackage{booktabs} 
\usepackage{enumerate}
\usepackage{subcaption}  
\usepackage{caption}
\usepackage{amsmath}
\usepackage{amsfonts}
\usepackage{multirow}
\usepackage{array}
\usepackage{makecell}
\usepackage{xspace}
\usepackage{adjustbox}
\usepackage{xcolor}
\usepackage{enumitem}

\usepackage{hyperref}

\newcommand{\showcomments}{yes}
\newcommand\zz[1]{
    \ifthenelse{\equal{\showcomments}{yes}}{\textcolor{blue}{\small [zz: #1]}}{\ignorespaces}
}
\newcommand\zy[1]{
    \ifthenelse{\equal{\showcomments}{yes}}{\textcolor{purple}{\small [zy: #1]}}{\ignorespaces}
}
\newcommand\cd[1]{
    \ifthenelse{\equal{\showcomments}{yes}}{\textcolor{brown}{\small [cd: #1]}}{\ignorespaces}
}

\usepackage[accepted]{mlsys2025}

\newcommand{\PPG}{PP-GNN\xspace}
\newcommand{\PPGs}{PP-GNNs\xspace}
\newcommand{\MPG}{MP-GNN\xspace}
\newcommand{\MPGs}{MP-GNNs\xspace}
\newcommand{\product}{\textit{ogbn-products}\xspace}
\newcommand{\pokec}{\textit{pokec}\xspace}
\newcommand{\wiki}{\textit{wiki}\xspace}
\newcommand{\paper}{\textit{ogbn-papers100M}\xspace}
\newcommand{\igbm}{\textit{igb-medium}\xspace}
\newcommand{\igbl}{\textit{igb-large}\xspace}

\mlsystitlerunning{Graph Learning at Scale: Characterizing and Optimizing Pre-Propagation GNNs}

\begin{document}

\twocolumn[
\mlsystitle{Graph Learning at Scale: Characterizing and Optimizing Pre-Propagation GNNs}

\mlsyssetsymbol{fn}{*}

\begin{mlsysauthorlist}
\mlsysauthor{Zichao Yue}{cu}
\mlsysauthor{Chenhui Deng}{nv,fn}
\mlsysauthor{Zhiru Zhang}{cu}
\end{mlsysauthorlist}

\mlsysaffiliation{cu}{Cornell University, Ithaca, New York, USA}
\mlsysaffiliation{nv}{NVIDIA, USA; *Work was done at Cornell}

\mlsyscorrespondingauthor{Zichao Yue}{zy383@cornell.edu}
\mlsyskeywords{Machine Learning, MLSys}

\vskip 0.1in

\begin{abstract}

Graph neural networks (GNNs) are widely used for learning node embeddings in graphs, typically adopting a message-passing scheme. This approach, however, leads to the \emph{neighbor explosion} problem, with exponentially growing computational and memory demands as layers increase. Graph sampling has become the predominant method for scaling GNNs to large graphs, mitigating but not fully solving the issue.
Pre-propagation GNNs (\PPGs) represent a new class of models that decouple feature propagation from training through pre-processing, addressing neighbor explosion in theory. Yet, their practical advantages and system-level optimizations remain underexplored.
This paper provides a comprehensive characterization of \PPGs, comparing them with graph-sampling-based methods in training efficiency, scalability, and accuracy. While \PPGs achieve comparable accuracy, we identify data loading as the key bottleneck for training efficiency and input expansion as a major scalability challenge. To address these issues, we propose optimized data loading schemes and tailored training methods that improve \PPG training throughput by an average of 15$\times$ over the \PPG baselines, with speedup of up to 2 orders of magnitude compared to sampling-based GNNs on large graph benchmarks. Our implementation is publicly available at \url{https://github.com/cornell-zhang/preprop-gnn}.

\end{abstract}
]
\printAffiliationsAndNotice{}

\section{Introduction}

Message-passing-based graph neural networks (\MPGs) have become a cornerstone for graph representation learning, achieving success in various tasks like node classification \cite{velickovic2017gat, wu2023gamora, kipf2016gcn}, link prediction \cite{zhang2018link, schutt2017schnet}, and graph clustering \cite{zhang2019attributed, ying2018dippool, tsitsulin2023graph}. However, scaling MP-GNNs to large graphs remains a significant challenge.



\begin{figure*}[t]
    \centering
    \includegraphics[width=0.95\textwidth]{ 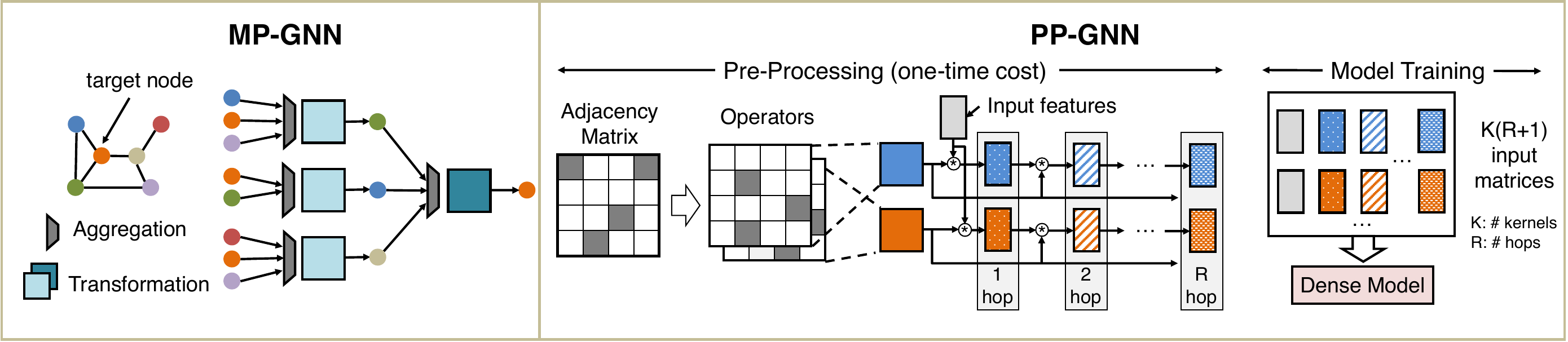}
    \vspace{-5pt}
    \caption{General structure of \MPG and \PPG models.}
    \label{fig:framework_overall}
    \vspace{-10pt}
\end{figure*}
The message-passing framework \cite{gilmer2017mpgnn} consists of two iterative steps: (1) feature aggregation and (2) transformation. Within this framework, each node collects feature embeddings from its neighbors and then transforms them using a learnable function. We show the architecture of \MPG models in Figure \ref{fig:framework_overall}. The main challenge in scaling \MPGs to large graphs stems from the ``neighbor explosion'' problem \cite{hamilton2017sage}, where nodes must recursively collect embeddings from increasingly larger neighborhoods across layers, causing the number of embeddings to grow exponentially with each additional layer. 

To address this challenge, 
various prior arts have introduced sampling-based GNNs to reduce the compute and memory footprint during message passing. Those models
encompass node-wise sampling to limit neighborhood sizes per node \cite{chen2017stochastic, hamilton2017sage}, layer-wise sampling to reduce node counts per layer \cite{chen2018fastgcn, zou2019ladies}, and graph-wise sampling to control overall subgraph size \cite{chiang2019cluster, zeng2019graphsaint}. 
However, the sampling-based GNNs face several major limitations. First, node-wise sampling methods only partially mitigate the neighbor explosion problem, as their time complexity still increases exponentially with the number of layers. 
More importantly, the sampling algorithms modify the graph topology by design, which inevitably breaks the functionality of computation graphs such as logic networks~\cite{wu2023gamora} and dataflow graphs~\cite{phothilimthana2024tpugraphs}, resulting in accuracy degradation on their downstream tasks~\cite{deng2024hoga}.
To circumvent the limitations of \MPGs, a new class of models known as pre-propagation GNNs (\PPGs) has emerged to tackle the scalability issue from a different angle \cite{wu2019SGC, frasca2020sign, dong2021pta, zhang2022gamlp, liao2022scara, chen2020gbp, deng2024hoga, zhu2020s2gc}. These models perform feature aggregation in a preprocessing step, eliminating the need for this computationally expensive step during model training.
This approach theoretically offers two advantages over \MPGs.
First, by decoupling nodes from interdependencies introduced by feature aggregation, 
nodes are processed independently during training,
effectively addressing the neighbor explosion problem. Second,
previous efforts \cite{gespmm} have shown that within the message-passing framework, feature aggregation is typically more time-consuming than transformation due to its sparse nature. By restricting training to dense computations, \PPGs are expected to achieve greater efficiency. Importantly, the input data preprocessing is a \textbf{one-time cost} that can be amortized across multiple rounds of hyperparameter tuning and model architecture adjustments. 

In this work, we conduct \textbf{the first systematic characterization of \PPG models}, comparing their training efficiency, scalability, and accuracy with \MPG models. We find that although \PPGs achieve comparable accuracy as \MPGs on commonly used large graph benchmarks, they do not exhibit clear training efficiency advantages over \MPGs that leverage tailored system-level optimizations. 

During our characterization, we identify two primary challenges for PP-GNN training: data loading as the major efficiency bottleneck and input expansion as the main scalability issue.
Data loading, consisting of batch assembly and data transfer, dominates training time due to lightweight computations in PP-GNNs, and the input expansion problem stems from the architecture of \PPG models, where the size of input features expands as more hops of neighbors are used, potentially exceeding host memory capacity.

To address these challenges, we propose several \textbf{system-level optimizations}. First, we introduce a customized data loader with an efficient batch assembly operation to reduce host-side data preparation overhead. Second, we design a double-buffer-based data prefetching scheme to decouple data loading from GPU-side computation. These optimizations significantly reduce data loading overhead while adhering to the standard training method, stochastic gradient descent with random reshuffling (SGD-RR) \cite{mishchenko2020sgdrr}.  Additionally, we propose chunk reshuffling, which shuffles training data at a coarser granularity, enabling bulk data transfer and more efficient GPU-side batch assembly. Moreover, we leverage chunk reshuffling to extend our training pipeline to leverage GPU direct storage (GDS) \cite{thompson2019gds} access, efficiently handling input sizes exceeding host memory capacity.

We further integrate these optimizations into an automated training configuration system for PP-GNNs, which detects hardware configurations and determines the best data placement and training strategies.
With these system optimizations, \PPGs demonstrate significantly higher training efficiency on various commonly used large graph benchmarks, including those with up to 100 million nodes.

Our main contributions are as follows:
\vspace{-10pt}
\begin{itemize}[left=0pt, labelsep=5pt]
    \setlength{\itemsep}{0pt plus 0.5pt}
    \setlength{\parsep}{0pt} 
    \setlength{\topsep}{0pt} 
    \setlength{\partopsep}{0pt} 
    \setlength{\parskip}{0pt} 
    \item We present the first comprehensive study comparing both the training efficiency and accuracy of \PPGs and \MPGs on commonly-used large graph benchmarks.
    \item We identify data loading as the critical efficiency bottleneck and the input expansion problem as the major scalability challenge for \PPG training.
    \item We propose various system-level optimizations to tackle the two challenges, including efficient batch assembly schemes, double-buffer-based data prefetching, and a tailored chunk reshuffling training method. Additionally, we propose an automated training configuration system for \PPGs that accommodates various graph sizes based on hardware configurations and a data placement policy. 
    \item Our optimizations improve \PPG training efficiency on average 15$\times$ compared to vanilla \PPG implementations and show comparable training efficiency when fetching input data directly from the solid-state drives (SSD) compared to from the host memory. After optimization, compared to \MPG models with state-of-the-art graph samplers, our optimized \PPGs achieve on average 9.9$\times$ and up to 2 orders of magnitude higher training throughput with higher accuracy on large graphs.
\end{itemize}
\section{Preliminaries}

This section introduces the key concepts related to GNNs, including the message-passing paradigm, graph sampling algorithms, and \PPGs. We also introduce existing \MPG training systems in this section. 

\subsection{Notations}
\label{subsec:notation}
In the following sections, we define a graph as $G = (V, E)$, where $V$ is the node set and $E$ is the edge set. We define $|V| = n$ as the total number of nodes in the graph and $|E| = m$ as the total number of edges. Each node $v \in V$ has a neighborhood set $N(v)$. Let $A \in \mathbb{R}^{n \times n}$ represent the adjacency matrix and $D \in \mathbb{R}^{n \times n}$ represent the diagonal degree matrix. $X \in \mathbb{R}^{n \times F}$ denotes the input node feature matrix with $F$ as the input feature dimension. 

\subsection{General Structure of \MPGs}

In general, GNN models take in graph-related information, such as the graph topology and node features, to learn latent node embeddings. Most popular GNN models can be generalized within the message-passing-based framework \cite{gilmer2017mpgnn}, which is articulated in Eq. \eqref{eq:generalstructure}. 
Here, $h_v^{(k)}$ denotes the embedding of node $v$ in layer $k$, with $h_v^{(0)} = x_v$. $l_k$ represents a transformation function, $f_k$ an aggregation function, and $e_{vu}$ the edge attribute between nodes $v$ and $u$.
\begingroup
\footnotesize
\begin{equation}
    h_v^{(k)} = l_k \left( h_v^{(k-1)},f_k \left( \{(h_u^{(k-1)}, e_{vu}) \mid \forall u \in N(v) \} \right) \right)
    \label{eq:generalstructure}
\end{equation}
\endgroup
Message passing occurs within the aggregation function $f_k$, where node $v$ aggregates embeddings from its one-hop neighbors, a process also known as feature propagation. As indicated in Equation \eqref{eq:generalstructure}, the aggregation function is applied recursively across layers, which can lead to the neighbor explosion problem. 
Various GNN models are specific instances of this generalized framework. For example, GraphSAGE \cite{hamilton2017sage} employs mean, Long Short-Term Memory (LSTM), or pooling aggregators in its aggregation function. The Graph Attention Network (GAT) \cite{velickovic2017gat} incorporates a learnable attention mechanism to assign different weights to neighbors during feature propagation. Both models employ a Multi-Layer Perceptron (MLP) as the transformation function.



\subsection{Graph Sampling}
\label{subsec:graphsampling}
Graph sampling is a widely adopted technique to scale \MPGs on large graphs, categorized into three types: node-wise sampling \cite{chen2017stochastic, hamilton2017sage, balin2024labor}, layer-wise sampling \cite{chen2018fastgcn, zou2019ladies}, and graph-wise sampling \cite{chiang2019cluster, zeng2019graphsaint}. 
Node-wise samplers, such as the one introduced in GraphSAGE \cite{hamilton2017sage}, limit neighborhood size during sampling but still face the neighbor explosion problem, with node count growing exponentially by layer. Layer-wise sampling methods sample a fixed number of nodes per layer, resulting in linear growth but struggling with sparse connectivity. LADIES \cite{zou2019ladies} tackles this problem with layer-dependent sampling for better connectivity. Graph-wise sampling methods, such as GraphSAINT \cite{zeng2019graphsaint}, sample subgraphs with a fixed number of nodes or edges, maintaining a subgraph size independent of model depth while ensuring connection.
LABOR \cite{balin2024labor} is a State-of-The-Art (SoTA) hybrid sampler combining the strength of both node-wise and layer-wise sampling, leading to fewer nodes sampled compared to node-wise samplers while maintaining an adaptive nature to different graph sizes.


\subsection{GNN Training Systems}


In sampling-based GNN training, the primary bottleneck is the graph sampling process, which includes node sampling and feature extraction \cite{liu2023bgl, yang2022gnnlab, lin2020pagraph}. Various training systems have been developed to optimize this process by leveraging GPUs. For instance, DGL \cite{wang2019dgl} accelerates node sampling and feature extraction on GPUs, provided the graph data fits entirely into GPU memory. PaGraph \cite{lin2020pagraph} utilizes GPU-based caching of node features while relying on CPU for node sampling. GNNLab \cite{yang2022gnnlab} employs GPUs for both node sampling and feature caching. These techniques can enhance \MPG training efficiency in both single-GPU and multi-GPU environments. Additional strategies employed in these systems include the use of NVLinks between GPUs to minimize communication overhead \cite{cai2023dsp}, hardware-aware graph partitioning \cite{sun2023legion,quiver2023}, and GPU kernel optimizaitons \cite{huang2024wisegraph, wang2019dgl}.


\subsection{Pre-propgation GNNs}

\PPGs \cite{frasca2020sign,deng2024hoga,zhang2022gamlp,chen2020gbp,liao2024ld2,liao2022scara,yu2020nars, wu2019SGC} have recently emerged as a promising approach to scaling GNN training. We show the general structure of \PPG models in Figure \ref{fig:framework_overall}. During preprocessing, node features are aggregated in a manner similar to feature propagation in \MPGs, but instead of relying directly on the adjacency matrix, operators derived from the adjacency matrix are typically employed. From a spectral perspective, these operators act as graph signal filters applied to the input graph \cite{gasteiger2019diffusion}. Given that most \MPGs effectively perform low-pass filtering on the input graph signal \cite{nt2019lowpass}, \PPGs can achieve comparable accuracy by learning on already filtered graph data. Like \MPGs, node features are propagated by multiplying the operators with the node feature matrix, yielding features at different hops through successive multiplications. The resulting node features are then stored and reused in the training phase, where a dense model is typically employed to learn node representations.

\PPG models can be generalized as follows:
\begingroup
\footnotesize
\begin{equation}
    \begin{aligned}
        \text{Preprocess: } 
        S_k &= \{ \mathbf{X}, \mathbf{B}_k \mathbf{X}, \ldots , \mathbf{B}_k^R \mathbf{X} \}, \quad k = 1, \ldots, K
    \end{aligned}
    \label{preprocess}
\end{equation}
\vspace{-10pt}
\begin{equation}
    \begin{aligned}
        \text{Train: } H &= l(S_1, \ldots , S_K),  
          \quad\quad Y = o(H)
    \end{aligned}
    \label{eq:train}
\end{equation}
\endgroup
\vspace{-10pt}   

In Eq. \eqref{preprocess}, X denotes the input feature matrix, R represents the number of hops, and $\mathbf{B}_k$ for $ k = 1, \ldots, K$ are $K$ operators. After preprocessing, we get $K$ sets of node features, denoted as $S_k$, each of which consists of $R+1$ matrices, corresponding to features that incorporate information from up to $R$-hop neighbors, along with the original node features. In Eq. \eqref{eq:train}, $l(\cdot)$ represents a specific learnable transformation function applied on all sets of $\mathbf{S}$, which outputs a single node embedding matrix $H$. An output function $o(\cdot)$ then transforms $H$ to the desired output shape.


For SGC \cite{wu2019SGC}, the operator is the normalized adjacency matrix expressed as $\mathbf{B}=\tilde{\mathbf{D}}^{-\frac{1}{2}} \tilde{\mathbf{A}} \tilde{\mathbf{D}}^{-\frac{1}{2}}$, where $\tilde{A} = I + A$ is the adjacency matrix with self-loops, and $\tilde{D}$ is the corresponding diagonal degree matrix. $o(\cdot)$ is a linear transformation while $l(\cdot)$ can be simplified as 
$\delta_{ir} = 1 \text{ if } i = r \text{ otherwise } 0$.
For SIGN \cite{frasca2020sign}, the operator can be the normalized adjacency matrix or those derived from Personalized PageRank (PPR) or Heat kernel \cite{gasteiger2019diffusion}.
The transformation function $l(\cdot)$ first concatenates matrices belonging to the same hop from different operators and then learns $R+1$ weight matrices for each hop. The output function $o(\cdot)$ is an MLP. 
For HOGA \cite{deng2024hoga}, the operator is captured by the normalized adjacency matrix. The transformation function adopts a transformer-like mechanism, which treats the $R+1$ input feature vectors of each node as $R+1$ tokens. An MLP is employed as the output function.


\section{\PPG Characterization}

As an emerging family of GNN models, \PPGs have yet to benefit from tailored system-level optimizations, raising questions about their practical training efficiency and scalability compared to \MPGs. Their model expressivity on commonly-used graph datasets has not been thoroughly explored either. In this section, we present a systematic characterization of \PPGs, examining their theoretical complexity, accuracy, training efficiency, and scalability.

\subsection{Complexity Analysis}
\label{subsec:motivation}
\begin{table}[!ht] 
\centering
\caption{Comparison of computational cost and memory complexity among GNN models --- Asymptotic complexities are shown without Big O notation. For computational cost, red denotes feature propagation, and blue denotes feature transformation.}
\begin{adjustbox}{max width=\columnwidth}
\begin{tabular}{l|l l}
\hline
\textbf{Model} & \textbf{Training Memory} & \textbf{Computational Cost} \\ \hline
GraphSAGE & $LbC^LF + LF^2$ & $\textcolor{red}{LFnC^{L+1}} + \textcolor{blue}{LnC^LF^2}$ \\
LADIES & $L^2bF + LF^2$ & $\textcolor{red}{L^2nFb} + \textcolor{blue}{L^2nF^2}$ \\
GraphSAINT & $LbF + LF^2$ & $\textcolor{red}{LnFb} + \textcolor{blue}{LnF^2}$ \\
LABOR & $LbC^LF + LF^2$ & $\textcolor{red}{LFnC^{L+1}} + \textcolor{blue}{LnC^LF^2}$\\ 
\hline 
SGC & $bF + F^2$ & $\textcolor{blue}{nF^2}$ \\
SIGN & $LbF + LF^2$ & $\textcolor{blue}{LnF^2}$ \\
HOGA & $LbF + LF^2 + Lb(r+1)^2$ & $\textcolor{blue}{Ln(r+1)F^2 + LnF(r+1)^2}$\\ 
\hline
\end{tabular}
\end{adjustbox}
\label{table:complexity_concise}
\vspace{-5pt}
\end{table}
First, we compare the theoretical computational cost and memory complexity of training \PPGs and \MPGs, as listed in Table \ref{table:complexity_concise}. 
In this table, we do not consider the sampling process for \MPGs. Here $b$ represents the mini-batch size, and we assume the number of nodes sampled per layer in LADIES and per subgraph in GraphSAINT is the same as $b$; $C$ represents the neighborhood size per node after sampling in GraphSAGE and LABOR, which is usually much smaller than $b$. To simplify the analysis, we assume the same dimension for the input layer and hidden layers, using $F$ to denote both. Other notations used in the complexity analysis are defined in Section~\ref{subsec:notation}.


There are two major components in the computational cost for \MPGs, one arising from feature propagation, denoted in red, and the other from feature transformation, denoted in blue. Prior studies \cite{gespmm} suggest that the former usually takes longer in practice due to its sparse nature. According to the table, \PPGs are expected to significantly boost the training efficiency, as they eliminate feature propagation from the training process. 

\subsection{Accuracy}

Conventional wisdom suggests that more scalable graph learning approaches like \PPGs may compromise accuracy for improved scalability. However, the learning capabilities of {\PPGs} are still not well understood and remain an active area of research \cite{chen2020graphaugmentedmlp}. Meanwhile, work from Deng et al. \cite{deng2024hoga} shows that \PPGs outperform \MPGs on several electronic design automation tasks that require complete neighbor information to infer functionality correctly. To this end, we investigate the model accuracy of various approaches on widely used large graph datasets, with detailed information in Table \ref{tab:datasets}. 

\begin{figure}[htp!]
    \vspace{-5pt}
    \centering
    \includegraphics[width=\columnwidth]{ 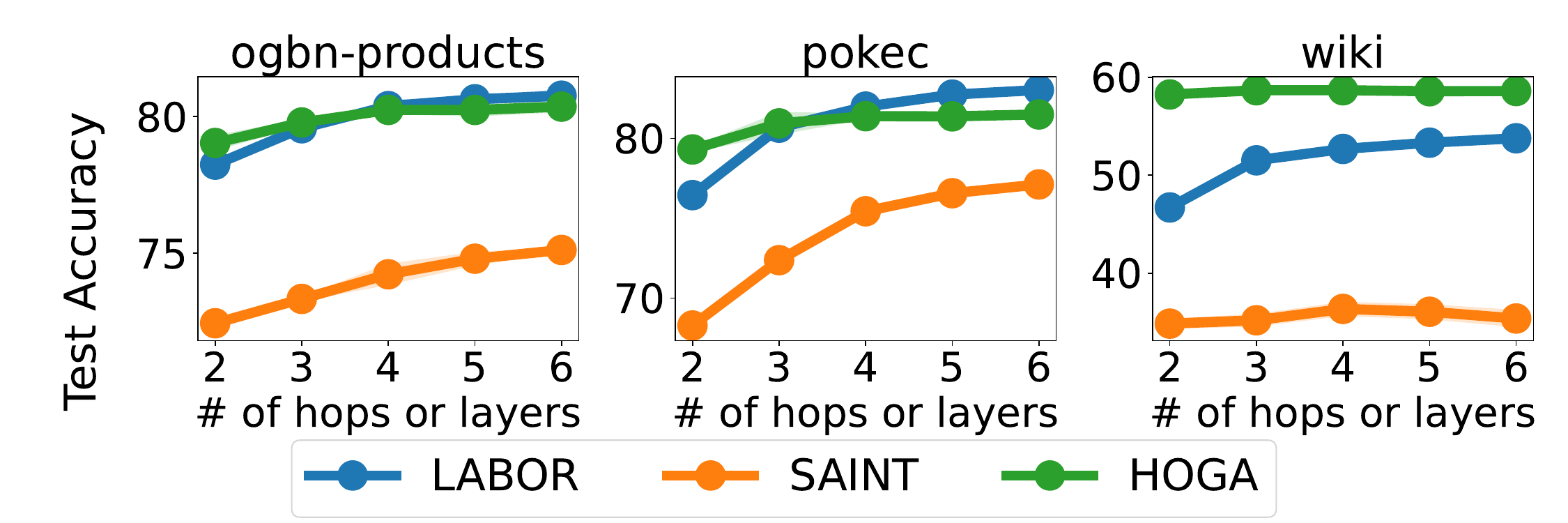}
    \vspace{-20pt}
    \caption{Test accuracy of GNN models with different hop counts or layer counts --- LABOR and SAINT represent GraphSAGE with the LABOR sampler and GraphSAINT node sampler, respectively.}
    \label{fig:accuracy}
    \vspace{-5pt}
\end{figure}

Our evaluation indicates that, among sampling methods, the LABOR sampler achieves the highest test accuracy across most settings, while HOGA outperforms other \PPG models in the majority of cases.
Figure \ref{fig:accuracy} illustrates the test accuracy of GraphSAGE \cite{hamilton2017sage} employing two different samplers—LABOR \cite{balin2024labor} and GraphSAINT \cite{zeng2019graphsaint}—as well as a \PPG model, HOGA, across varying numbers of layers or hops on three datasets. Complete results and hyperparameter settings are provided in Appendix~\ref{appendix-parameter} and \ref{appendix-preto}.
From Figure \ref{fig:accuracy} we observe two trends: (1) \textbf{\PPGs demonstrate accuracy comparable to \MPGs}, with LABOR serving as a representative; (2) \textbf{Expanding the node receptive field tends to improve test accuracy on large graphs}---unlike prior arts evaluated on small graphs, using 5/6 layers (hops) further improve accuracy in our experiments, which is consistent with the trend seen in \cite{chiang2019cluster}. 


\subsection{Practial Training Efficiency}
\label{subsec:praticaltraingefficiency}
To evaluate the practical training efficiency, we compare \MPGs and \PPGs from two perspectives: convergence rate and epoch time.
\begin{figure}[ht]
    \centering
    \includegraphics[width=1.0\columnwidth]{ 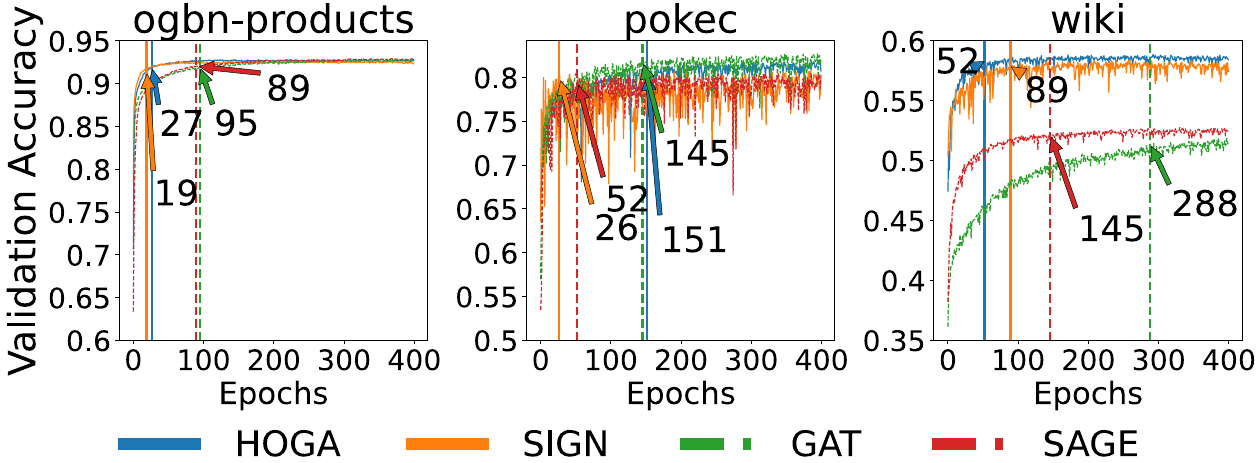} 
    \vspace{-7pt}
    \caption{Convergence rate comparison among 4-layer(hop) \MPGs and \PPGs --- The number in the plot denotes the convergence point where 99\% of peak validation accuracy is reached.}
    \label{fig:convergence}
    \vspace{-5pt}
\end{figure}

\textbf{Convergence Rate Comparison.}
The convergence rate significantly impacts the end-to-end training time of a GNN model. Since different models reach different peak accuracies even on the same dataset, to make a fair comparison, we measure the convergence point as the epoch where each model first reaches 99\% of its peak validation accuracy. Figure \ref{fig:convergence} shows the convergence points of different \MPGs and \PPGs with hyperparameters tuned for optimal accuracy, detailed in Appendix \ref{appendix-parameter}. Complete results are provided in Appendix \ref{appendix-converge}. Our results show that \PPGs have on par or faster convergence rate than \MPGs.

\textbf{Epoch Time Comparison.}
For \PPGs, we implemented SIGN \cite{frasca2020sign}, HOGA \cite{deng2024hoga}, and SGC \cite{wu2019SGC} in PyTorch, which use the PyTorch data loader to decouple data preparation from model training. For comparison, we also implemented \MPG models with various graph sampling algorithms in DGL \cite{wang2019dgl}, adhering to the structures described in their original papers \cite{yingpinsage, velickovic2017gat, zou2019ladies, balin2024labor, zeng2019graphsaint}. Detailed settings are provided in Appendix~\ref{appendix-parameter}.
With DGL version 0.8 and later, the graph sampling process can be GPU-accelerated. When input data is pinned in host memory, DGL utilizes NVIDIA's UVA \cite{schroederuva} technology, 
allowing GPU to access data on the host memory with zero-copy. Preloading the input data to GPU memory further reduces end-to-end training time due to the high memory bandwidth available on GPU.

\begin{figure}[ht]
    \centering
    \includegraphics[width=0.9\columnwidth]{ 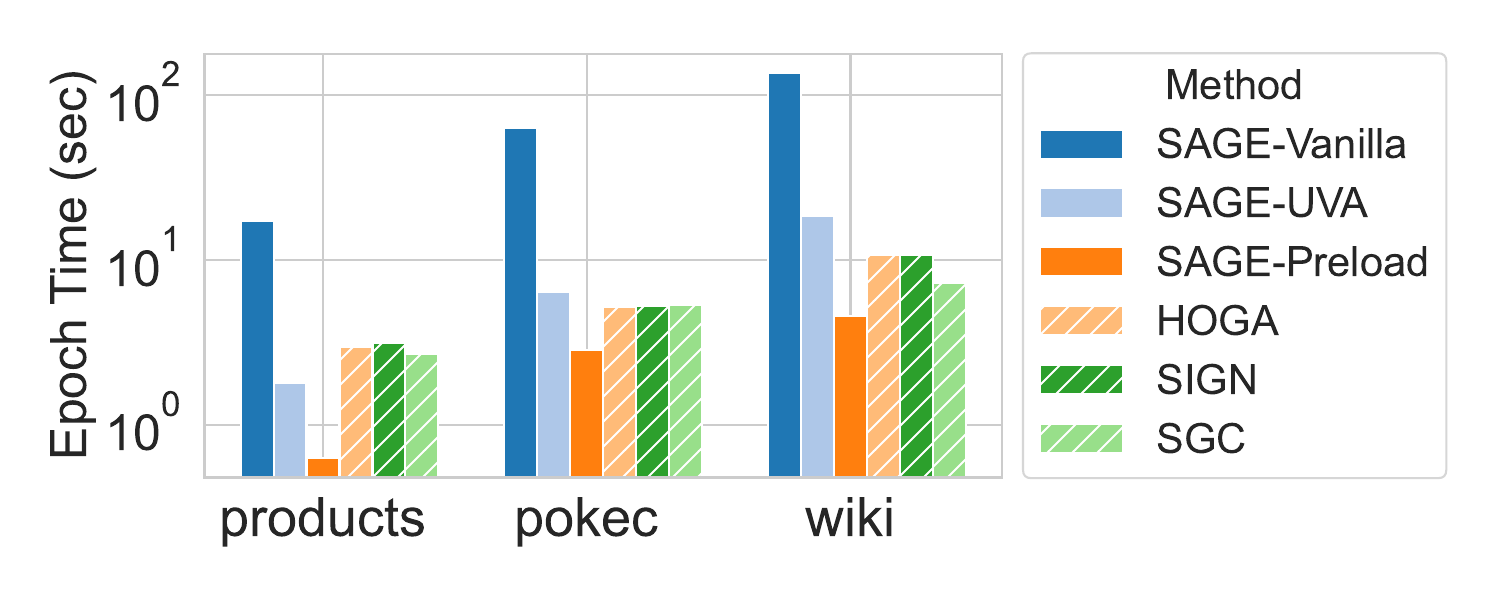} 
    \vspace{-10pt}
    \caption{Epoch time comparison of \PPGs and \MPGs --- \PPGs (HOGA, SIGN, and SGC) are shown with diagonal hatching, while SAGE represents GraphSAGE with the LABOR sampler, with UVA and preload indicating DGL optimizations.}
    \label{fig:training_time_nonopt}
\end{figure}
Figure \ref{fig:training_time_nonopt} compares the epoch times among 3-layer \MPGs (GraphSAGE with the LABOR sampler) and 3-hop \PPGs (SIGN, HOGA, and SGC). The epoch time for GraphSAGE is measured with a vanilla DGL implementation with CPU-based graph sampling and two optimized versions, denoted as SAGE-UVA and SAGE-Preload,  representing GPU-based graph sampling with the use of UVA and input preloading, respectively. From Figure \ref{fig:training_time_nonopt} we see system-level optimizations significantly improve MP-GNN training throughput. Consequently, despite theoretical advantages in computational cost (Table \ref{table:complexity_concise}), vanilla \PPG implementations take longer epoch time than \MPGs fully optimized in DGL.

\begin{figure}[ht]
    \centering
    \includegraphics[width=0.9\columnwidth]{ 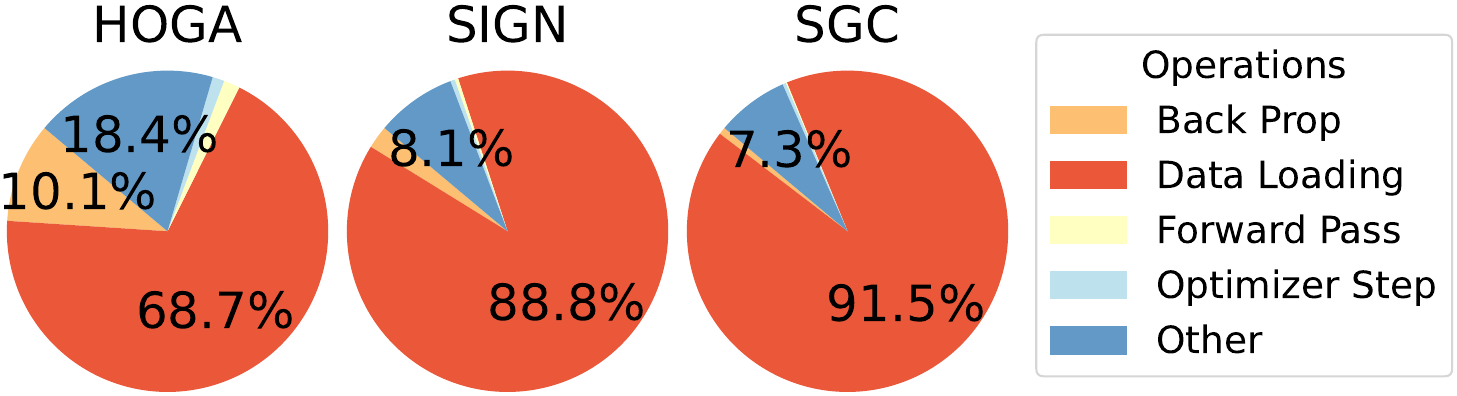} 
    \vspace{-10pt}
    \caption{Training time breakdown of \PPGs on \product.}
    \label{fig:training_time_decomp}
\end{figure}
Further investigation reveals that the primary overhead in the baseline implementation stems from data loading. Figure \ref{fig:training_time_decomp} shows the epoch time breakdown of three \PPG models on the \product dataset, averaged across various hops. The figure highlights that computation is relatively lightweight for \PPGs, which is represented as the forward pass, backpropagation, and optimizer step in the figure. In contrast, the epoch time is dominated by data loading. Therefore, system-level optimizations that improve data loading efficiency are crucial for \PPGs to achieve their potential training efficiency advantages.


\subsection{The Challenge of Input Expansion}
Through our characterization, we identify a largely overlooked issue in prior studies of \PPGs, which we term the ``input expansion problem.'' 
From Eq. \eqref{eq:train} we observe that $K(R+1)$ input matrices are generated during preprocessing, each representing an input feature matrix with $i\in R$ hop neighbor information from one of $K$ operators. 
Consequently, the input feature 
size is expanded to $K(R+1)$ times larger, as illustrated in Figure \ref{fig:framework_overall} (d). 
For instance, the \igbl dataset \cite{khatua2023igb} takes 400 GB for the input features. With a small $R$ and $K$, like $R=3$ and $K=1$, the input data for \PPGs will expand to 1.6 TB, exceeding the typical host memory capacity. 
Consequently, randomly fetching features from the storage system during training will result in severe training efficiency degradation due to low storage random read speed. Therefore, we need system-level solutions to overcome the input expansion problem of \PPGs on large graphs. 

\subsection{Pre-processing Overhead}
Compared to \MPGs, \PPGs require an additional preprocessing step. However, this preprocessing can be considered a one-time cost, as the processed data is stored and reused throughout the training process. Table \ref{tab:datasets} presents the preprocessing time for the datasets used in our evaluation. Typically, training a GNN model involves hundreds of epochs per run, and hyper-parameter tuning may require tens or even hundreds of such runs. As shown in Table \ref{tab:datasets}, the preprocessing overhead is usually much smaller than the time required for a single training run, and thus, can be efficiently amortized over the entire training phase. 
\section{System Optimizations}
\begin{figure*}[ht]
    \centering
    \includegraphics[width=0.95\textwidth]{ 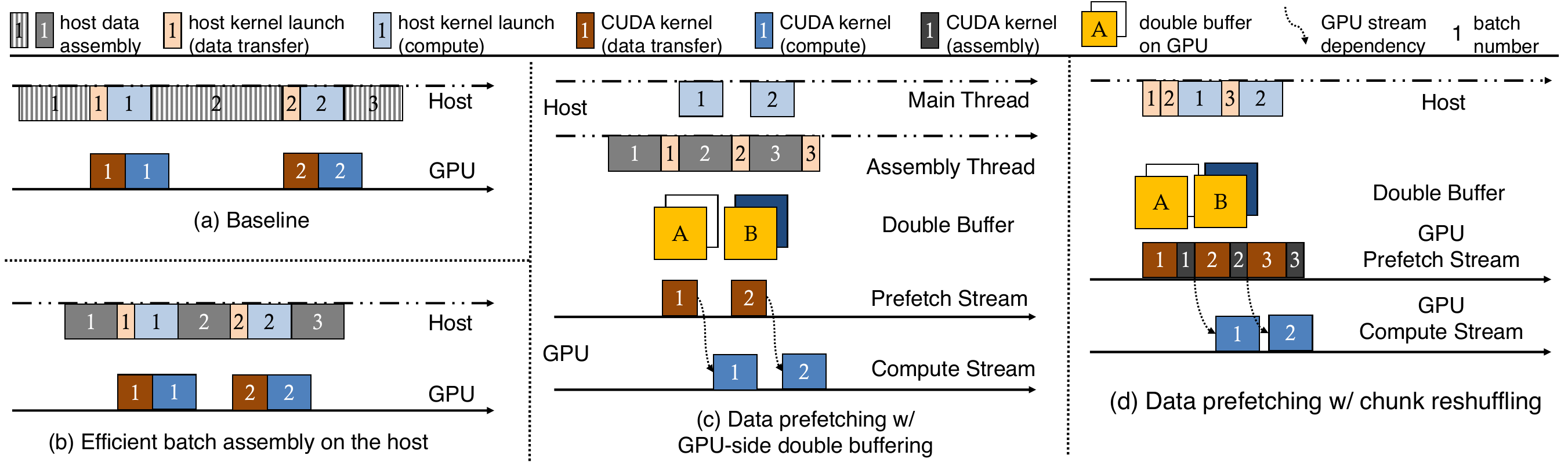} 
    \vspace{-5pt}
    \caption{System-level optimizations adopted in our work --- For the case with input data residing in the host memory.}
    \label{fig:system_op_overall}
    \vspace{-10pt}
\end{figure*}
Our characterization reveals that data loading time significantly dominates the training time of \PPGs. Typically, data loading consists of two steps: batch assembly and data transfer. During batch assembly, the data loader extracts node vectors belonging to the current batch, and transfers them to GPU in the following data transfer step. To reduce the data loading overhead, a straightforward solution involves loading input data into GPU memory to leverage its high bandwidth. However, the input expansion problem limits the feasibility of this approach, as GPU memory typically has a much lower capacity compared to host memory.


To reduce the data loading overhead while maintaining the input data in host memory, we propose several strategies. First, we devise a custom data loader with efficient data indexing to reduce the kernel launching overhead during batch assembly (Section~\ref{subsec:customizedloader}). Second, we introduce a double-buffer-based data prefetching mechanism on GPU, which largely hides data loading time by pipelining it with computation (Section~\ref{subsec:customizedloader}). Last, we develop a chunk reshuffling method that allows us to reorder batch assembly and data transfer, enabling GPU-side batch assembly, taking advantage of high GPU memory bandwidth (Section~\ref{subsec:chunkreshuffling}).
Moreover, chunk reshuffling paves the way for scaling to large graphs. By replacing host memory access with GPU direct storage access, we can easily handle input sizes exceeding the host memory capacity (Section~\ref{subsec:directstorageaccess}).    
\subsection{Customized Data Loading}
\label{subsec:customizedloader}

Upon profiling the \PPG baseline implementations, we observe that the PyTorch data loader extracts node features individually during batch assembly, resulting in frequent kernel invocations on the host side. As a result, batch assembly dominates the total training time, as depicted in Figure \ref{fig:system_op_overall} (a).
To mitigate the redundant kernel launching, we design a customized data loader utilizing the \texttt{index} operator provided by PyTorch to copy the scattered node features into a pinned tensor in host memory, which is then transferred to the GPU asynchronously. This approach is feasible due to the simplicity of the input data format, as \PPG inputs are purely dense tensors. By launching the \texttt{index} operator only once per batch, we significantly reduce the kernel launching overhead, as illustrated in Figure \ref{fig:system_op_overall} (b).

Despite this improvement, batch assembly on the host side still incurs significant time, potentially exceeding GPU computation time. This is primarily due to the extraction of scattered data in memory, limited by the host memory bandwidth. A potential solution is to cache node features on GPU to leverage its high memory bandwidth, as adopted by many \MPG systems \cite{yang2022gnnlab, sun2023legion}. However, this approach is unsuitable for \PPGs, as the training data lacks both temporal and spatial locality, being accessed only once in a random order every epoch.
Instead, we implement a data prefetching scheme using double buffers on GPU, as shown in Figure \ref{fig:system_op_overall} (c). This approach decouples data loading from GPU-side computation, enabling pipelining of these two steps. To achieve this, we use separate threads on the host side for launching compute kernels and data-loading-related kernels. On the GPU side, different streams are utilized for data prefetching and computation. 
As shown in Figure \ref{fig:system_op_overall} (c), our prefetching scheme effectively hides the batch assembly overhead.

\subsection{Chunk Reshuffling}
\label{subsec:chunkreshuffling}
While double-buffer-based prefetching pipelines data loading with computation, it fails to fully eliminate overhead when data loading time exceeds computation time. This overhead arises from (1) batch assembly, constrained by host memory bandwidth, and (2) data transfer, limited by the host-to-GPU interconnect.  Since data transfer is already optimized by Direct Memory Access (DMA) technique, further reducing its duration is challenging.

To reduce batch assembly time, we propose a chunk reshuffling training method. In this method, at the start of each epoch, we reshuffle training data indices at the chunk level, with each chunk comprising contiguous node features. Then, we transfer individual chunks belonging to the current batch from host memory to GPU and assemble chunks into batches. This approach takes advantage of the significantly higher DRAM bandwidth on GPU for batch assembly. Although data transfer overhead increases as more DMA transfer kernels are launched, this is minor provided the chunk size is sufficiently large. The efficacy of this approach is illustrated in Figure \ref{fig:system_op_overall} (d).
 Lastly, chunk reshuffling can be considered a form of insufficient shuffling scheme \cite{meng2019insufficientshuffle, nguyen2022localshuffle}, which is commonly used in practice. In Section \ref{subsec:chunk_reshuffle}, our empirical results show that chunk reshuffling has negligible impacts on test accuracy and convergence rate for \PPGs.

 \subsection{Direct Storage Access}
 \label{subsec:directstorageaccess}

The input expansion problem can cause the preprocessed input data of large graphs to exceed the host memory capacity. A na\"{\i}ve implementation fetching individual node features from storage to the host would suffer from slow random reads, resulting in significant data loading overhead.

Our chunk reshuffling method provides a foundation for extending to storage-based training; reading chunks from the storage system is significantly more efficient compared to reading individual node features. By replacing host-memory-reading operators with GPU direct storage access, we retain the benefits of pipelined data loading and computation with our double-buffer-based data prefetching scheme.
In our implementation, we leverage the NVIDIA GDS technique \cite{thompson2019gds} for direct storage access, which automatically utilizes DMA engines and system buffers, ensuring efficient data transfers under various system configurations. 
To maximize the parallel processing capabilities of modern storage systems and bus bandwidth utilization, we split input features of different hops into separate files, enabling parallel storage access requests.
\section{Automated Training Configuration}
Building on our system-level optimizations, we extend our training pipeline to develop an automated configuration system tailored for \PPGs. This system automates key configurations, particularly for data placement and training methods, optimizing \PPG training based on hardware resources and model characteristics. Implemented in PyTorch, our system offers a user-friendly interface, allowing integration of \PPG models without model-specific system tweaking. 
Before training starts, our system assesses the available hardware resources, including the number of GPUs and GPU and host memory capacities. To determine the minimum GPU memory space requirement for a specific model, we adopt an approach similar to PaGraph \cite{lin2020pagraph}, where we conduct a one-time training session using storage-based data loading to measure peak GPU memory usage. Combining the information of input data size, our configuration system automatically decides data placement and corresponding training method, as detailed below.

\textbf{GPU memory.} 
 Preloading input data to GPU memory is prioritized due to its high bandwidth. For large datasets, our system supports distributing data across multiple GPUs, with the data loader fetching data in a locality-aware manner \cite{yang2019localoading} to adapt to SGD-RR. When data is preloaded to GPU memory, our double-buffer-based prefetching further enhances training efficiency. However, with the high bandwidth of GPU memory, batch assembly is not a bottleneck, making chunk reshuffling negligible for performance. Thus, SGD-RR is preferred in this scenario.
 
\textbf{Host Memory.} 
When the input data exceeds GPU memory capacity, it is placed in host memory. With chunk reshuffling, the entire input data must be pinned in host memory for non-blocking transfers. Otherwise, only a buffer proportional to the mini-batch size is pinned. The configure system defaults to SGD-RR for large data to avoid excessive host memory pinning, unless specified by users.

\textbf{Storage.} When the input data exceeds the host memory capacity, our system allows the GPU to fetch data directly from the storage via NVIDIA GDS \cite{thompson2019gds}. Currently, we only support chunk reshuffling in this scenario, since SGD-RR requires fine-grained data access, significantly increasing data loading overhead. 

The influence of data placement on training throughput is evaluated in Appendix \ref{appendix-placement}.

\section{Evaluation}

\label{subsec:experimentsetup}
\textbf{Experiment Setup.}
We conduct the experiments on a Linux server with two 3.0 GHz Intel Xeon CPUs, 380 GB RAM and four RTX A6000 GPUs. Detailed hardware and software configurations are provided in Appendix \ref{appendix-software}.

\textbf{Datasets.}
\begin{table*}[!ht]
\centering
\caption{Dataset statistics --- This table summarizes key dataset statistics, with the last column showing the one-time cost of pre-processing, as both wall-clock time and a proportion of a single training run, amortizable over multiple runs (details are provided in Appendix \ref{appendix-preprocessing}).}
\begin{adjustbox}{max width=\textwidth}
\begin{tabular}{c c c c c c c c c}
\hline
Dataset & \#Nodes & \#Edges & \%Labeled & \makecell{Split \\ (train/valid/test)} & \makecell{\#Input \\ Features} & \#Classes & \makecell{Size  \\ (graph/node)} & \makecell{ Pre-processing Time / sec \\ (compare to single training run)} \\ \hline
ogbn-products & 2,449,029 & 61,859,140 & 100\% & 0.08/0.02/0.9 & 100 & 47 & 0.9 GB/0.9 GB & 51.8 (53\%)  \\ \hline
pokec & 1,632,803 & 30,622,564 & 100\% & 0.5/0.25/0.25 & 65 & 2 & 0.5 GB/0.4 GB & 27.59 (3\%) \\ \hline
wiki & 1,925,342 & 303,434,860 & 100\% & 0.5/0.25/0.25 & 600 & 5 & 4.5 GB/4.3 GB & 122.79 (11\%)\\ \hline
IGB-medium & 10,000,000 & 120,077,694 & 100\% & 0.6/0.2/0.2 & 1024 & 19 & 1.8 GB/39.0 GB & 386.63 (11\%) \\ \hline
ogbn-papers100M & 111,059,956 & 1,615,685,872 & 1.4\% & 0.78/0.08/0.14 & 128 & 172 & 24 GB/53 GB & 507.8 (90\%) \\ \hline
IGB-large & 100,000,000 & 1,223,571,364 & 100\% & 0.6/0.2/0.2 & 1024 & 19 & 19 GB/400 GB & 4521.5 (28\%)\\ \hline
\end{tabular}
\label{tab:datasets}
\end{adjustbox}
\vspace{-5pt}
\end{table*}
We use three medium-sized graphs, \product, \pokec, and \wiki, each with approximately 2 million nodes for a detailed investigation of the accuracy-efficiency trade-off between \PPGs and \MPGs. While we classify these as medium-sized, it's important to note that these graphs are usually considered large graphs in the GNN community \cite{chiang2019cluster}. We use three large graphs, \textit{ognb-papers100M}, \textit{IGB-meduim}, and \textit{IGB-large} to investigate the scalability of \PPGs under different scenarios. The datasets are chosen from real-world benchmarks including Open Graph Benchmark (OGB) \cite{hu2020ogb}, Illinois Graph Benchmark (IGB) \cite{khatua2023igb}, and the Cornell non-homophilous benchmark \cite{lim2021hetero}. The dataset attributes are listed in table \ref{tab:datasets}. Our paper focuses on node classification tasks, which serve as the foundation for link and graph classification.

\textbf{\MPG.}
We use GraphSAGE \cite{hamilton2017sage} and GAT \cite{velickovic2017gat} as backbone models. We employ the samplers from GraphSAGE \cite{hamilton2017sage}, LABOR \cite{balin2024labor}, LADIES \cite{zou2019ladies} and GraphSAINT \cite{zeng2019graphsaint}
 and refer to them as Neighbor, LABOR, LADIES, and SAINT in the following sections, respectively. GraphSAGE is set with a hidden dimension of 256, using the mean aggregator, and GAT is set with a hidden dimension of 128 per channel across 4 channels. We adopt two commonly used 3-layer fanout settings: [15 10 5] for GraphSAEG and [10 10 10] for GAT. This configuration pushes GAT towards accuracy and GraphSAGE towards efficiency, in line with the model complexity and hidden dimension setting, and offers a balanced view of the accuracy-efficiency trade-off of \MPG models. Detailed hyperparameter settings are provided in Appendix \ref{appendix-parameter}.

\textbf{\PPG.}
We choose SGC \cite{wu2019SGC}, SIGN \cite{frasca2020sign}, and HOGA \cite{deng2024hoga} as our \PPG models. SGC represents the simplest form of \PPGs, consisting of only one linear layer and using only one input matrix. HOGA, which adopts a transformer-like multi-head attention scheme, is a relatively complex \PPG model with higher expressivity, while SIGN lies in between. For these \PPGs, we use a single operator: the normalized adjacency matrix. For SIGN, we use 3 layers with a hidden dimension of 512. For HOGA, we use a hidden dimension of 256 for medium-sized graphs and 1024 for large graphs, with a single multi-head attention layer. This configuration pushes HOGA towards accuracy and SIGN towards efficiency. Detailed hyperparameter settings are provided in Appendix \ref{appendix-parameter}.

\textbf{Baselines.}
Our \PPG baselines are implemented in PyTorch, leveraging PyTorch \texttt{DataLoader} for data loading. 
The \texttt{pin\_memory} attribute is enabled in \texttt{DataLoader} and 2 workers are used to achieve optimal performance. 
For \MPGs, we implement them in DGL \cite{wang2019dgl}, GNNLab \cite{yang2022gnnlab}, SALIENT++ \cite{SALIENT++} and Ginex \cite{park2022ginex}.

\subsection{Accuracy-Efficiency Comparison}
\begin{figure}[ht]
    \centering
    \includegraphics[width=0.98\columnwidth]{ 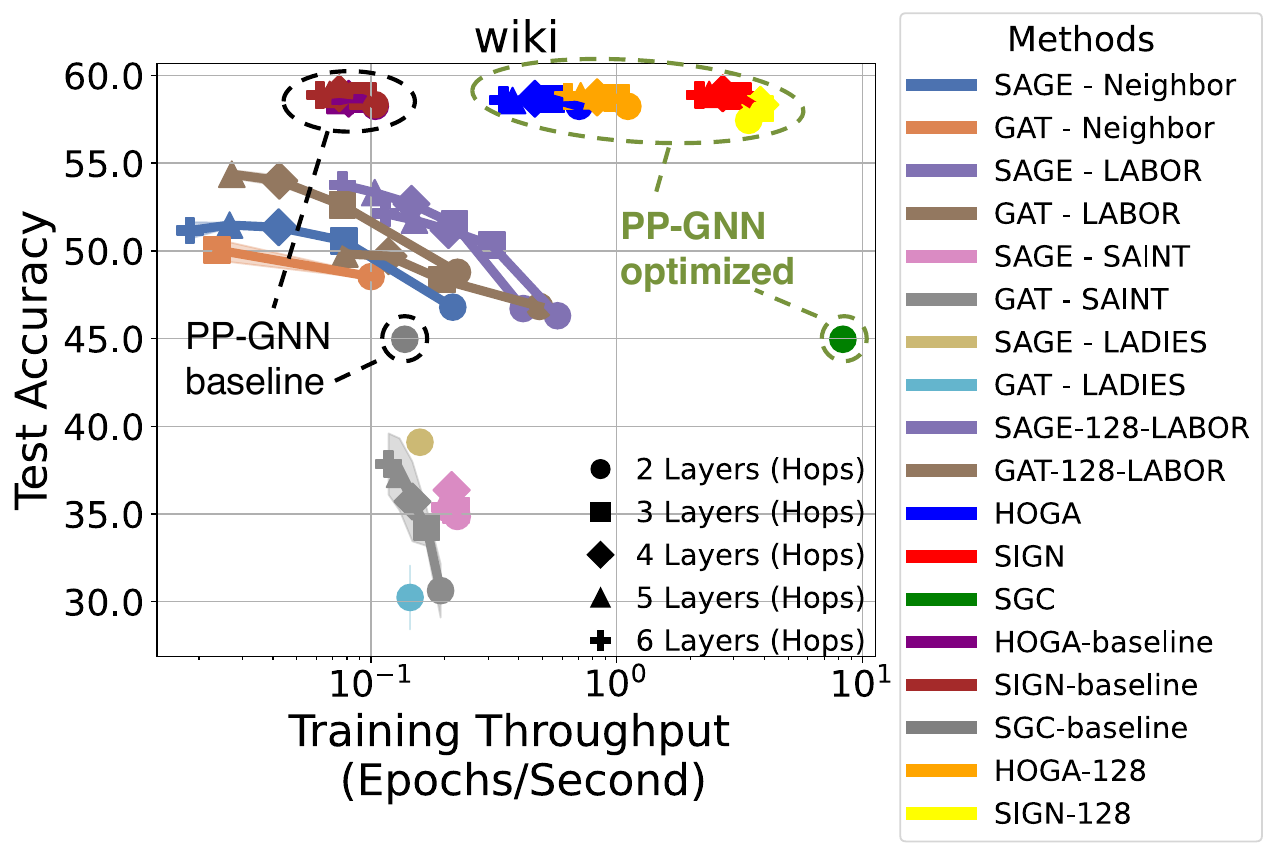} 
    \vspace{-5pt}
    \caption{Accuracy efficiency trade-off comparison among \MPGs and \PPGs on \wiki --- The \MPG legend keys show the backbone model and the graph sampler. '128' in the labels represents the additional hidden dimension setting.}
    \label{fig:Paretowiki}
    \vspace{-15pt}
\end{figure}
This section compares the accuracy and training efficiency of \PPGs and \MPGs after applying our proposed system-level optimizations. Figure \ref{fig:Paretowiki} shows the accuracy-efficiency trade-offs of \PPGs and \MPGs on the \wiki dataset, with the Pareto frontier indicated in the upper right direction. Diagrams for the \product and \pokec datasets are shown in Appendix \ref{appendix-preto} (Figure \ref{fig:pareto_frontier_products_pokec}). Hyperparameter settings are detailed in Appendix \ref{appendix-parameter}. 
The primary variable in these experiments is the node receptive field, defined as the number of layers in \MPGs and the number of hops in \PPGs. All experiments are conducted on a single GPU with data preloaded into GPU memory. GNNLab's GPU-side input caching does not outperform DGL preloading, and we present only DGL results.

Figure \ref{fig:Paretowiki} and Figure \ref{fig:pareto_frontier_products_pokec} demonstrate that our optimizations push the \PPGs to the Pareto frontier on all three datasets. It also shows that HOGA and SIGN achieve comparable accuracy to \MPGs with node-wise samplers, with significantly higher training efficiency. SGC, while fastest among all approaches, sacrifices substantial accuracy due to not fully utilizing all the hops.
Among all the graph sampling approaches, the SoTA LABOR sampler achieves the Pareto-optimum but still suffers from the neighbor explosion problem to some extent. LADIES and SAINT overcome this problem with a significant sacrifice in test accuracy, occupying the lower parts in the figures.

An important trend observed is that a larger node receptive field enhances model accuracy (a pattern true for both \MPGs and \PPGs, though the accuracy gains are more pronounced in \MPGs). Given that the training time of \PPGs increases sub-linearly with additional hops, these models become increasingly competitive in terms of training efficiency as the node receptive field expands. For example, on \wiki, SIGN is 9 times faster than SAGE-LABOR with 2 layers or hops, and this advantage grows to 28 times with 6 layers or hops. Moreover, we employ a smaller hidden dimension setting, 128, across models besides their original settings. Reducing the hidden dimension to 128, \MPGs experience up to 4\% accuracy loss, while HOGA and SIGN only see a 0.5\% variation. With a smaller hidden dimension of 128, SIGN is 136$\times$ faster than GAT with a hidden dimension of 512 with 5 hops or layers, while maintaining an accuracy advantage of 3.9\% on the test set.


\subsection{Influence of Chunk Reshuffling}
\label{subsec:chunk_reshuffle}
\begin{figure}[ht]
    \centering
    \includegraphics[width=0.9\columnwidth]{ 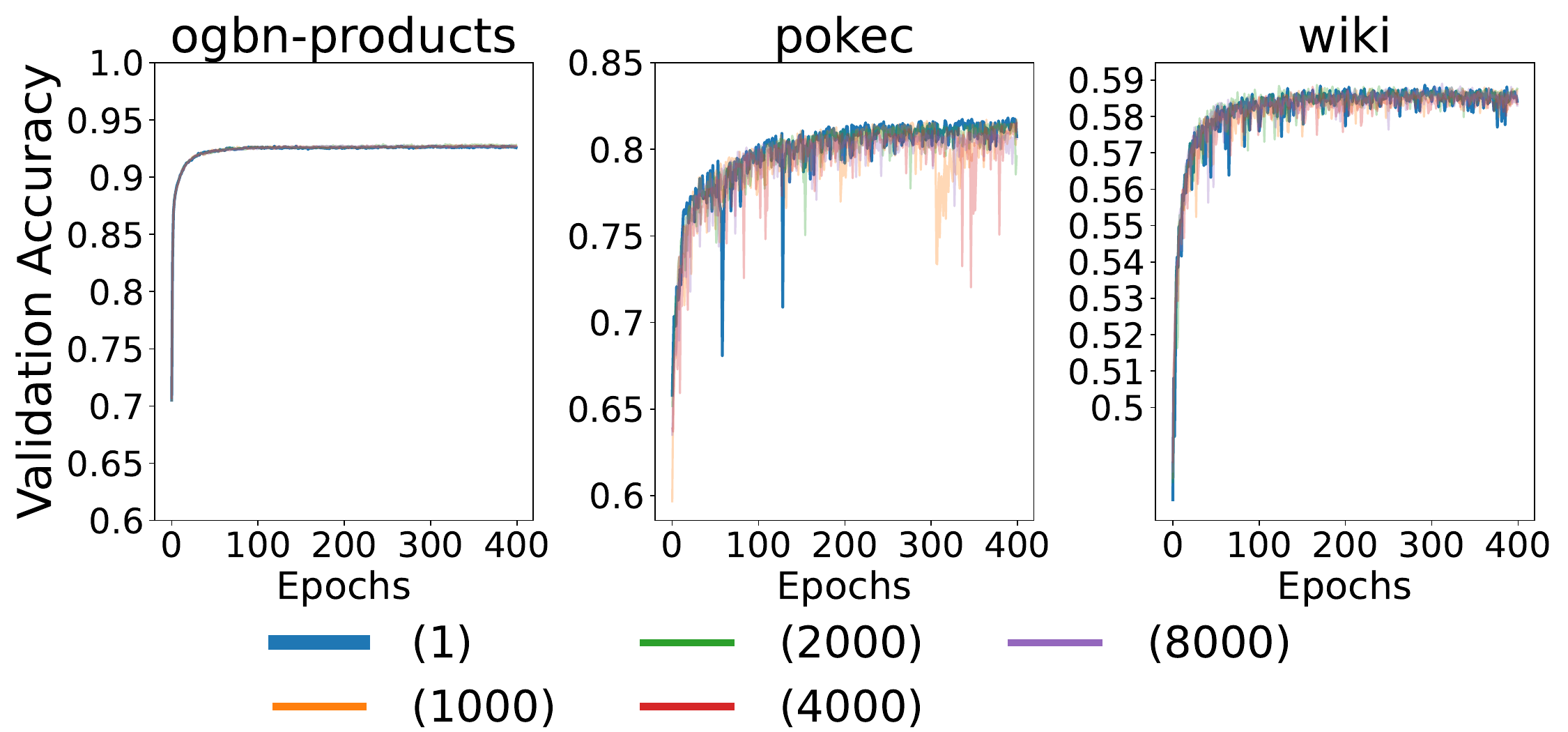} 
    \vspace{-10pt}
    \caption{Validation accuracy of HOGA with 4 hops on three datasets --- The number in the legend denotes the chunk size.}
    \label{fig:SGDCR_HOGA}
\end{figure}
In this section, we investigate the impact of our proposed chunk reshuffling training method on model accuracy and convergence rate using the three medium-sized graphs. For \PPG models, we fix all the hyperparameters except the chunk size, which is selected from {1, 1000, 2000, 4000, 8000}. Figure \ref{fig:SGDCR_HOGA} shows the validation accuracy of HOGA with 4 hops. From the figure, we observe that for \product, the variation in the training curves across different chunk sizes is negligible, with mean test accuracy varying by less than 0.1\%. For \pokec and \wiki, although there are some fluctuations in the training curves, the final test accuracy difference is less than 0.5\%. This trend is consistent across other \PPG models and hop settings, with complete results in Appendix \ref{appendix-chunkreshuffle}. Consequently, in the following sections, we use a chunk size of 8000, equal to the batch size, when employing the chunk reshuffling training method. 

\subsection{Ablation Study}
\begin{figure}[ht]
    \centering
    \includegraphics[width=0.9\columnwidth]{ 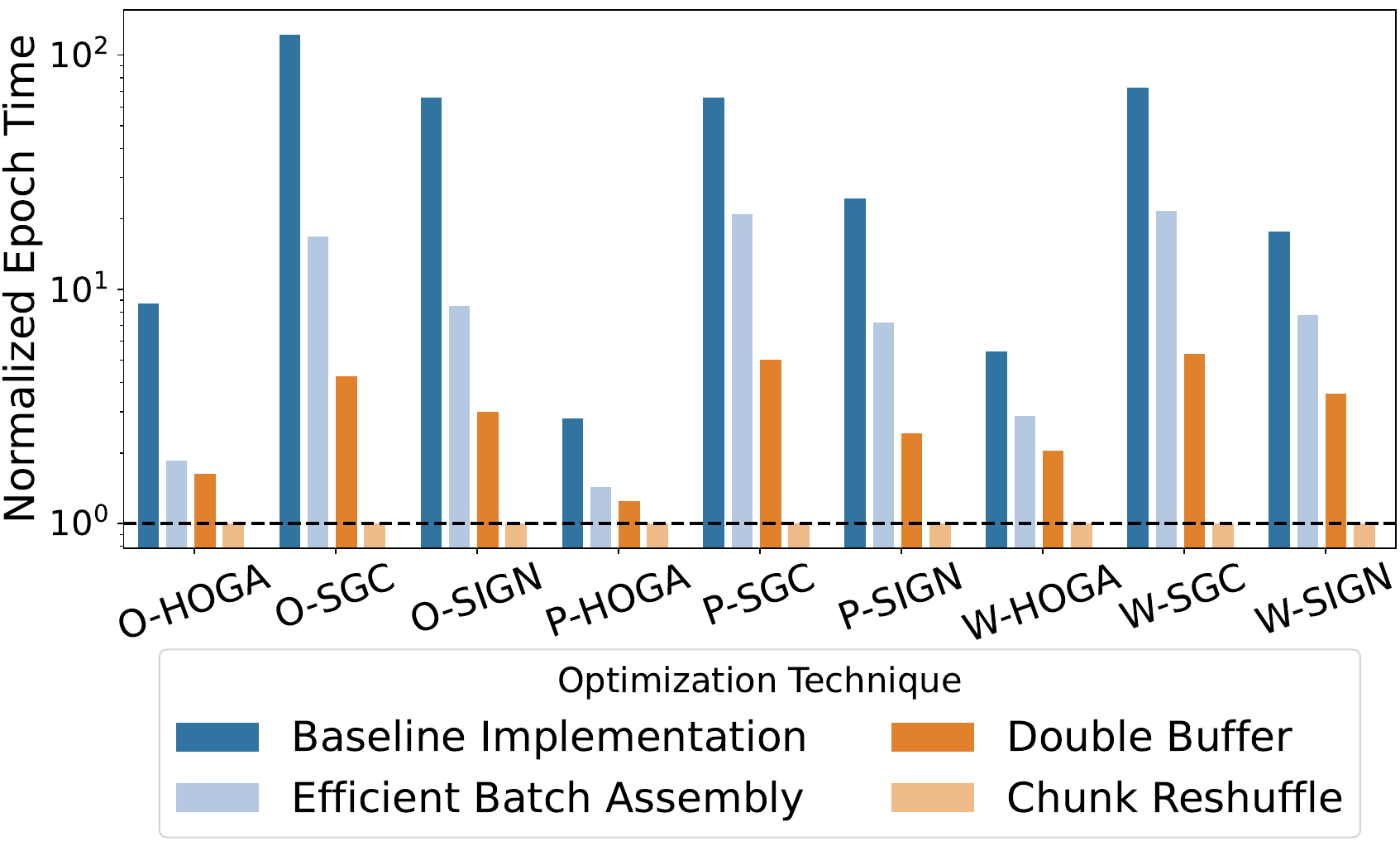}
    \vspace{-10pt}
    \caption{Ablation study with input data in the host memory --- X-ticks show the dataset (O: \product, P: \pokec, W: \wiki) and the \PPG model (HOGA, SIGN and SGC).}
    \label{fig:ablation}
    \vspace{-7pt}
\end{figure}

In this section, we evaluate the efficacy of our techniques for improving data loading efficiency: efficient host-side batch assembly, GPU-side double-buffer-based prefetching, and chunk reshuffling with GPU-based data assembly. Figure \ref{fig:ablation} presents the normalized epoch time of various \PPG models across different datasets, with data averaged over 2 to 6 hops and 100 epochs using the geometric mean.

Our results show that efficient host-side batch assembly provides a 3.3$\times$ speedup over the baseline implementation. Adding double-buffer-based prefetching yields an additional 1.9$\times$ speedup. Further, chunk reshuffling with GPU-based batch assembly delivers an additional 2.4$\times$ speedup, resulting in a total 15$\times$ improvement over the baseline. More detailed analysis can be referred to Appendix \ref{appendix-ablationstudy}.

\subsection{Results on Large Graphs}
\label{subsec:largegraph}
We examine the training efficiency and scalability of different methods on three large graph datasets using GraphSAGE as the \MPG model, while HOGA and SIGN as the \PPG models. In DGL, we adopt the LABOR sampler, while in GNNLab, SALIENT++, and Ginex, we use their hardcoded neighbor samplers respectively. Inference of GNNLab relies on an older version of DGL which CUDA12.1 does not support, hence we do not report the test accuracy of GNNLab.

\begin{table}[ht]
\centering
\small
\caption{Performance comparison on \paper, with test accuracy averaged over 5 runs of 100 epochs each.}
\resizebox{\columnwidth}{!}{
\begin{tabular}{ccccccc}
\toprule[1pt]
\multirow{2}{*}{\textbf{\makecell{Layers\\or Hops}}} & \multirow{2}{*}{\textbf{Model}} & \multirow{2}{*}{\textbf{\makecell{Training\\System}}} & \multirow{2}{*}{\textbf{\makecell{Test\\Acc (\%)}}} & \multicolumn{3}{c}{\textbf{Throughput (epoch/sec)}} \\
\cmidrule(lr){5-7}
& & & & \textbf{1 GPU} & \textbf{2 GPUs} & \textbf{4 GPUs} \\
\midrule
\multirow{4}{*}{2} 
    & \multirow{2}{*}{SAGE} & DGL    & 64.43$\pm$0.19 & 0.12 & - & - \\ 
    &                       & SALIENT++ & 64.28$\pm$0.16             & 0.27 & 0.46 & 0.42 \\ 
    &                       & GNNLab & -             & 0.72 & 0.71 & 1.33 \\ \cline{2-7}
    & SIGN                  & Ours   & 65.70$\pm$0.09 & 2.94 & 3.23 & 6.62 \\ 
    & HOGA                  & Ours   & \textbf{66.19$\pm$0.08} & 0.53 & 0.77 & 1.54 \\ 
\midrule[1pt]
\multirow{4}{*}{3} 
    & \multirow{2}{*}{SAGE} & DGL    & 65.79$\pm$0.14 & 0.04 & - & - \\ 
    &                       & SALIENT++ & 65.67$\pm$0.06             & 0.05 & 0.10 & 0.10 \\
    &                       & GNNLab & -             & 0.19 & 0.19 & 0.29 \\ \cline{2-7}
    & SIGN                  & Ours   & 66.29$\pm$0.20 & 2.92 & 3.22 & 6.54 \\ 
    & HOGA                  & Ours   & \textbf{66.65$\pm$0.09} & 0.41 & 0.61 & 1.23 \\
\midrule[1pt]
\multirow{4}{*}{4} 
    & \multirow{2}{*}{SAGE} & DGL    & 66.44$\pm$0.17 & 0.02 & - & - \\ 
    &                       & SALIENT++ & 65.78$\pm$0.12             & 0.01 & 0.03 & 0.04 \\
    &                       & GNNLab & -             & 0.06 & 0.07 & 0.10 \\ \cline{2-7}
    & SIGN                  & Ours   & 66.36$\pm$0.05 & 2.86 & 3.13 & 6.25 \\ 
    & HOGA                  & Ours   & \textbf{66.86$\pm$0.11} & 0.36 & 0.54 & 1.09 \\ 
\bottomrule[1pt]
\end{tabular}
}

\label{tab:papers100M}
\end{table}

First, the \paper graph dataset features instances where \textbf{labeled nodes constitute only a minor portion of the total node count}. For \PPGs, the input data size after preprocessing is proportional to the number of labeled nodes, while the information of unlabeled nodes is incorporated during preprocessing. Notably, the original input features for \paper occupy 53 GB, but the labeled part only takes 0.8 GB per hop after preprocessing, fitting comfortably into GPU memory. Conversely, \MPGs require accessing the entire graph topology and all input features during training, totaling 77 GB, which exceeds a single GPU’s capacity, making loading all input data into GPU memory infeasible. For \MPGs, we use DGL-UVA, SALIENT++, and GNNlab to evaluate their training efficiency on the \paper dataset.

Table \ref{tab:papers100M} shows training throughput and test accuracy under 100 epochs for different approaches with 2 to 4 hops or layers. HOGA achieves the highest accuracy among all methods, with up to 1.76\% higher accuracy than SAGE. DGL achieves higher accuracy than SALIENT++ due to the adoption of the LABOR sampler.
In terms of training efficiency, SIGN and HOGA achieve up to 5$\times$ and 41$\times$ higher throughput than GraphSAGE on a single GPU. Compared to DGL-UVA, GNNLab improves training efficiency by caching input features and graph topology on GPU, but its hardcoded graph sampler produces larger subgraphs than LABOR, offsetting its caching benefits as the number of layers increases. Due to the large graph size, we encounter out-of-memory (OOM) issues when extending the DGL-UVA to multiple GPUs. We employ SALIENT++ and GNNLab in the scalability study. \PPGs achieve higher scalability than \MPGs implemented in both SALIENT++ and GNNLab. One exception is SALIENT++ with 4 layers. Under this setting, SALIENT++ encounters OOM issues with a batch size of 8000, and we need to reduce the batch size to 1000. With a smaller batch size, SALIENT++ shows better scalability, with sacrifice on training throughput. Across all settings, \PPGs consistently outperform \MPGs with 4 GPUs, with up to 156$\times$ speedup. 

\begin{table}[ht]
\centering
\small
\caption{Performance comparison on \igbm, with test accuracy averaged over 5 runs of 20 epochs each.}
\resizebox{\columnwidth}{!}{
\begin{tabular}{ccccccc}
\toprule[1pt]
\multirow{2}{*}{\textbf{\makecell{Layers\\or Hops}}} & \multirow{2}{*}{\textbf{Model}} & \multirow{2}{*}{\textbf{\makecell{Training\\System}}} & \multirow{2}{*}{\textbf{\makecell{Test\\Acc (\%)}}} & \multicolumn{3}{c}{\textbf{Throughput (epoch/min)}} \\
\cmidrule(lr){5-7}
& & & & \textbf{1 GPU} & \textbf{2 GPUs} & \textbf{4 GPUs} \\
\midrule
\multirow{6}{*}{2} 
    & \multirow{2}{*}{SAGE} & DGL    & 75.44$\pm$0.02 & 0.35 & 0.39 & 0.77 \\ \cline{3-7}
    &                       & GNNLab & -             & 2.83 & 2.78 & 7.68 \\ \cline{2-7}
    & \multirow{2}{*}{SIGN}                 & Ours-RR    & 76.16$\pm$0.02 & 3.16 & 4.23 & 6.59 \\
    &                                       & Ours-CR    & \textbf{76.17$\pm$0.02} & 9.35 & 6.04 & 11.13 \\
    & \multirow{2}{*}{HOGA}                 & Ours-RR    & 76.08$\pm$0.03 & 2.22 & 4.03 & 5.99 \\
    &                                       & Ours-CR    & 76.07$\pm$0.03 & 5.43 & 4.11 & 7.85 \\ 
\midrule[1pt]
\multirow{3}{*}{3} 
    & SAGE & DGL    & 75.47$\pm$0.05 & 0.10 & 0.11 & 0.21 \\ \cline{2-7}
    & SIGN                  & Ours-RR   & \textbf{76.17$\pm$0.02} & 2.44 & 3.42 & 4.87 \\ 
    & HOGA                  & Ours-RR   & 76.10$\pm$0.03 & 1.65 & 2.93 & 4.49 \\
\bottomrule[1pt]
\end{tabular}
}
\vspace{-5pt}
\label{tab:IGBmedium_full}
\end{table}

We use the \igbm dataset to assess the training efficiency of \MPGs and \PPGs in scenarios where the \textbf {dataset size exceeds GPU memory capacity}. 
\igbm is fully labeled, with an input feature dimension of 1024, occupying 40 GB for input features, exceeding the single GPU memory capacity for both \MPGs and \PPGs.


Table \ref{tab:IGBmedium_full} presents the training throughput and test accuracy over 20 epochs for different approaches. \PPGs consistently achieve higher test accuracy than \MPGs on this dataset. In terms of training throughput, \PPGs with chunk reshuffling significantly outperform other methods, with up to 24$\times$ speedup compared to \MPGs with 3 hops or layers. GNNLab performs comparably to \PPGs with SDG-RR and outperforms DGL-UVA by a wide margin due to its GPU-side input feature caching, which mitigates the high data extraction and transformation demands of \igbm stemming from its large input feature dimension. However, with more than 2 layers, GNNLab encounters OOM issues from larger sampled subgraphs. 

When scaling to 4 GPUs, \PPGs with SGD-RR show similar scalability as \MPGs. Although \PPGs with chunk reshuffling achieve higher training efficiency, they demonstrate relatively less scalability, delivering only 1.27 $\times$ average speedup when using 4 GPUs, which is primarily bottlenecked by host-to-GPU bandwidth, and using more GPUs does not mitigate the problem. This issue is more pronounced with direct storage access, as storage systems typically have less bandwidth to the host or GPU. Therefore, we only implement single GPU direct storage access.


\begin{table}[ht]
\centering
\small
\caption{Performance comparison on \igbl, with test accuracy reported under 3 epochs.}
\resizebox{\columnwidth}{!}{
\begin{tabular}{ccccc}
\toprule[1pt]
\textbf{\makecell{Layers\\or Hops}} & \textbf{Model} & \textbf{\makecell{Training\\System}} & \textbf{\makecell{Test \\Accuracy (\%)}} & \textbf{\makecell{Throughput\\(epoch/hour)}} \\ \hline
\multirow{4}{*}{2} & \multirow{2}{*} {SAGE} & DGL & 63.07  & 0.77 \\ \cline{3-5}
& & Ginex & 63.09 & 0.65 \\ \cline{2-5}
& SIGN & Ours & 64.41 & 10.52 \\ 
& HOGA & Ours & \textbf{64.42} & 8.58 \\ 
\midrule[1pt]
\multirow{4}{*}{3} 
& \multirow{2}{*} {SAGE} & DGL & 62.85 & 0.17 \\ \cline{3-5}
& & Ginex & 62.73 & 0.19 \\ \cline{2-5}
& SIGN & Ours & 64.41 & 8.06 \\ 
& HOGA & Ours & \textbf{64.52} & 6.67 \\
\bottomrule[1pt]
\end{tabular}
}

\label{tab:igblarge}
\end{table}

Lastly, we utilize the \igbl dataset to demonstrate that our proposed optimizations can effectively address the input expansion problem when the \textbf{input data size exceeds the host memory capacity}. For \MPGs, we adopt two baselines, Ginex \cite{park2022ginex} and DGL. Ginex is a storage-based \MPG training system leveraging host-side caching. For DGL, we employ the \texttt{mmap} technique to map the input feature file into memory, which allows DGL to access necessary data portions directly from storage without loading the entire dataset into host memory. After preprocessing, the input data for \PPGs occupies approximately 1.6 TB with 1 kernel and 3 hops. Table \ref{tab:igblarge} presents the training throughput and test accuracy for different approaches. We limit the number of epochs to 3 and the number of runs to 1 due to the prolonged execution time of \MPGs, in line with the IGB official leaderboard. Our results reveal that \PPGs achieve up to 42$\times$ greater training throughput compared to \MPGs, highlighting the superior performance of \PPGs on ultra-large graphs. Conversely, the excessive training time per epoch renders detailed hyperparameter tuning impractical for \MPGs. 

Compared to GraphSAGE, HOGA, and SIGN achieve training throughput improvements of up to 2 orders of magnitude, with an average of 9.9$\times$ across three large graph datasets, while maintaining superior accuracy. These results make HOGA and SIGN compelling options for efficient learning on large graphs.

\section{Conclusions}
This work presents the first comprehensive study comparing the training efficiency and accuracy of \PPGs with \MPGs on large graph benchmarks. While \PPGs match \MPGs in accuracy, tailored system optimizations are crucial for realizing their theoretical efficiency and scalability. Our proposed optimizations help \PPGs achieve on average 9.9$\times$ higher training throughput on large graph datasets compared to \MPGs optimized in SOTA \MPG training systems while maintaining higher accuracy. 
\section{Acknowledgments}
This work was supported in part by ACE, one of the seven
centers in JUMP 2.0, a Semiconductor Research Corporation (SRC) program sponsored by DARPA and NSF Awards \#2212371 and \#2403135, and a Qualcomm Innovation Fellowship. We appreciate the input and discussions from Yixiao Du,
Yaohui Cai, Dr. Mingyu Liang, and the anonymous reviewers.

\bibliography{references}
\bibliographystyle{mlsys2025}

\appendix
\section{Hyperparameter Settings}
\label{appendix-parameter}
\textbf{\MPG.}
For backbone \MPG models, we use the DGL example implementations of GraphSAGE and GAT. GraphSAGE is set with a hidden dimension of 256, using the mean aggregator, and GAT is set with a hidden dimension of 128 per channel across 4 channels. 

For node-wise sampling methods, including Neighbor and LABOR samplers, we adopt two commonly used 3-layer fanout settings: [15 10 5] for GraphSAEG and [10 10 10] for GAT. Building on the 3-layer setup, we extend it to 4, 5, and 6 layers with smaller fanout limits to avoid OOM issues, using [15 10 5 3 3 3] for GraphSAGE and [10 10 10 5 5 5] for GAT. For 2-layer models, we adjust the fanout to [15 10] for GraphSAGE and [10 10] for GAT for consistency. This configuration pushes GAT towards accuracy and GraphSAGE towards efficiency, in line with the model complexity and hidden dimension setting, and offers a balanced view of the accuracy-efficiency trade-off of \MPG models. For LADIES, we set the nodes sampled per layer to 512, following the largest node limitation used in their original paper \cite{zou2019ladies}. For GraphSAINT, we use the node sampler and set the node limitation to the same as the batch size. 

Regarding batch size, the commonly used choices in the literature include 512, 1024, 2000, 4000, and 8000. A larger batch size helps reduce epoch time since the total number of sampled nodes is reduced, with increased memory requirement and generally requires more epochs to converge. In our experiments, we choose a batch size of 8000, which leads to a higher training throughput of \MPGs while permitting convergence under 400 epochs which is used as the total number of epochs per run for the medium-sized graphs. 

In our accuracy-efficiency trade-off exploration, we fine-tune two hyperparameters, the learning rate and dropout rate on all datasets except \igbl. The learning rate is chosen from [0.01, 0.001], and the dropout rate is chosen from [0.1, 0.2, 0.3, 0.4, 0.5, 0.6, 0.7].  Due to resource constraints, a more thorough investigation of hyperparameters is left for future work.

\textbf{\PPG.}
For \PPG models, we follow the implementations from their official GitHub repos. For SIGN, we use 3 layers with a hidden dimension of 512. For HOGA, we fine-tune the hidden dimension from two settings: 256 with 1 head or 64 with 4 heads, with a single multi-head attention layer. On the three large graphs, we use a hidden dimension of 256 with 4 heads instead. For all three models, we fine-tune the learning rate and dropout rate as for \PPG models, chosen from [0.01 0.001] and [0.1, 0.2, 0.3, 0.4, 0.5, 0.6, 0.7], respectively. For a fair comparison, we set the batch size the same as \MPGs to 8000. For operators, we use only one kernel, the normalized adjacency matrix, and choose between directed or undirected adjacency matrix depending on which yields higher accuracy.

\section{Convergence Comparison}
\label{appendix-converge}
\begin{figure*}[ht]
    \centering
    \begin{minipage}[b]{0.49\textwidth}
        \centering
        \includegraphics[width=\textwidth]{ 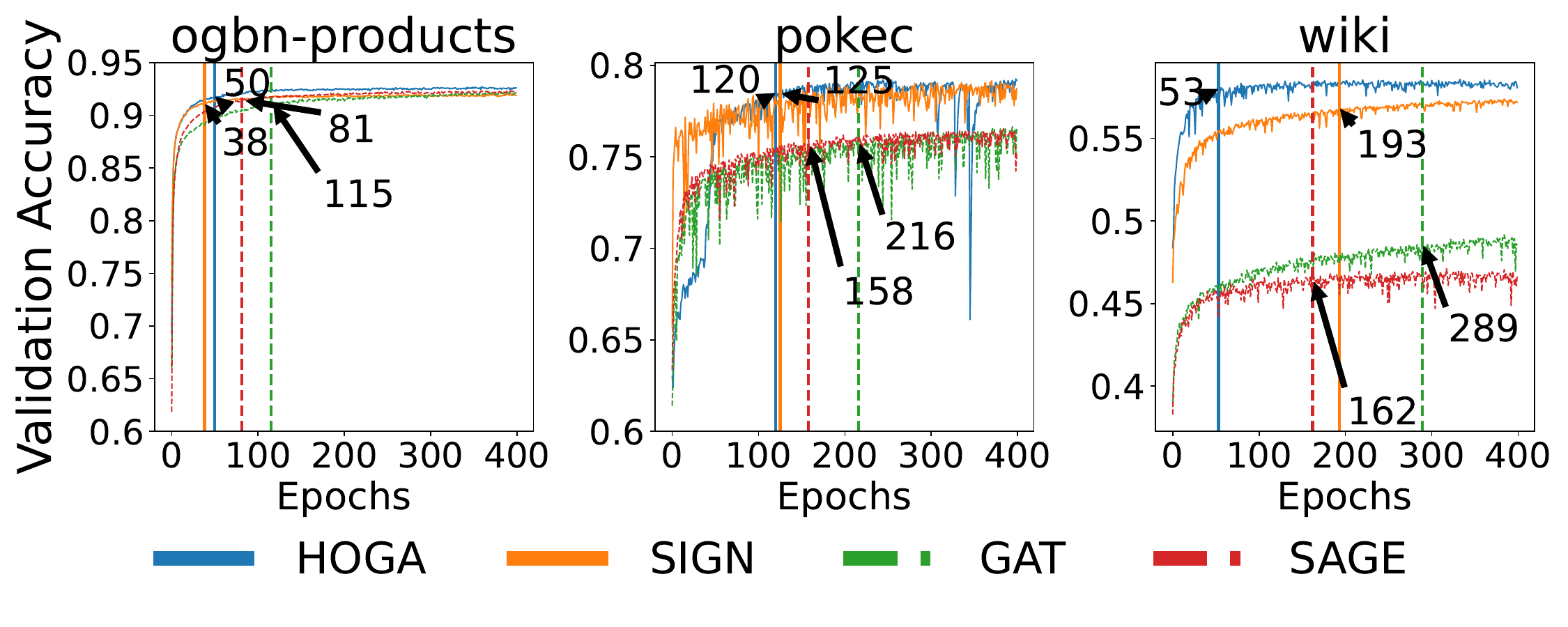}
        \subcaption{Convergence rate comparison under 2 layers or hops.}
        \label{fig:plotA_converge}
    \end{minipage}
    \hfill
    \begin{minipage}[b]{0.49\textwidth}
        \centering
        \includegraphics[width=\textwidth]{ 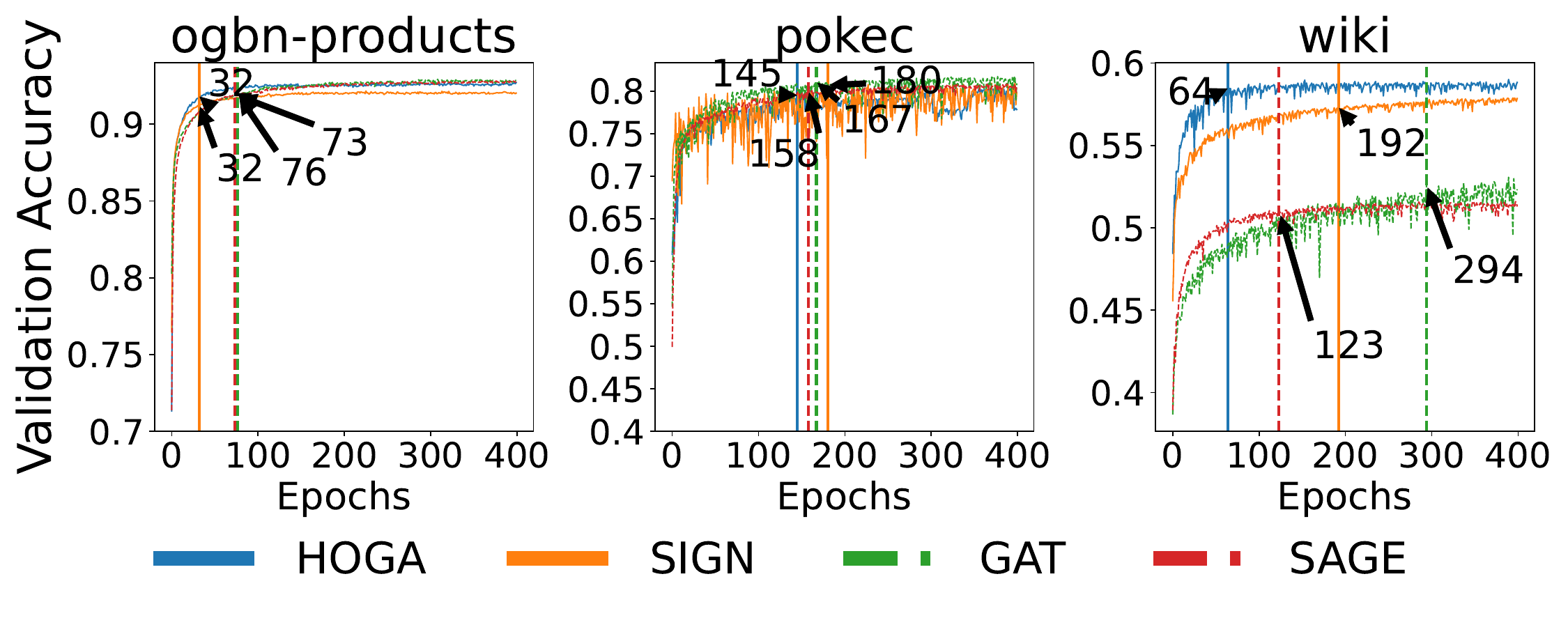}
        \subcaption{Convergence rate comparison under 3 layers or hops.}
        \label{fig:plotB_converge}
    \end{minipage}

    \vspace{0.5cm} 

    \begin{minipage}[b]{0.49\textwidth}
        \centering
        \includegraphics[width=\textwidth]{ 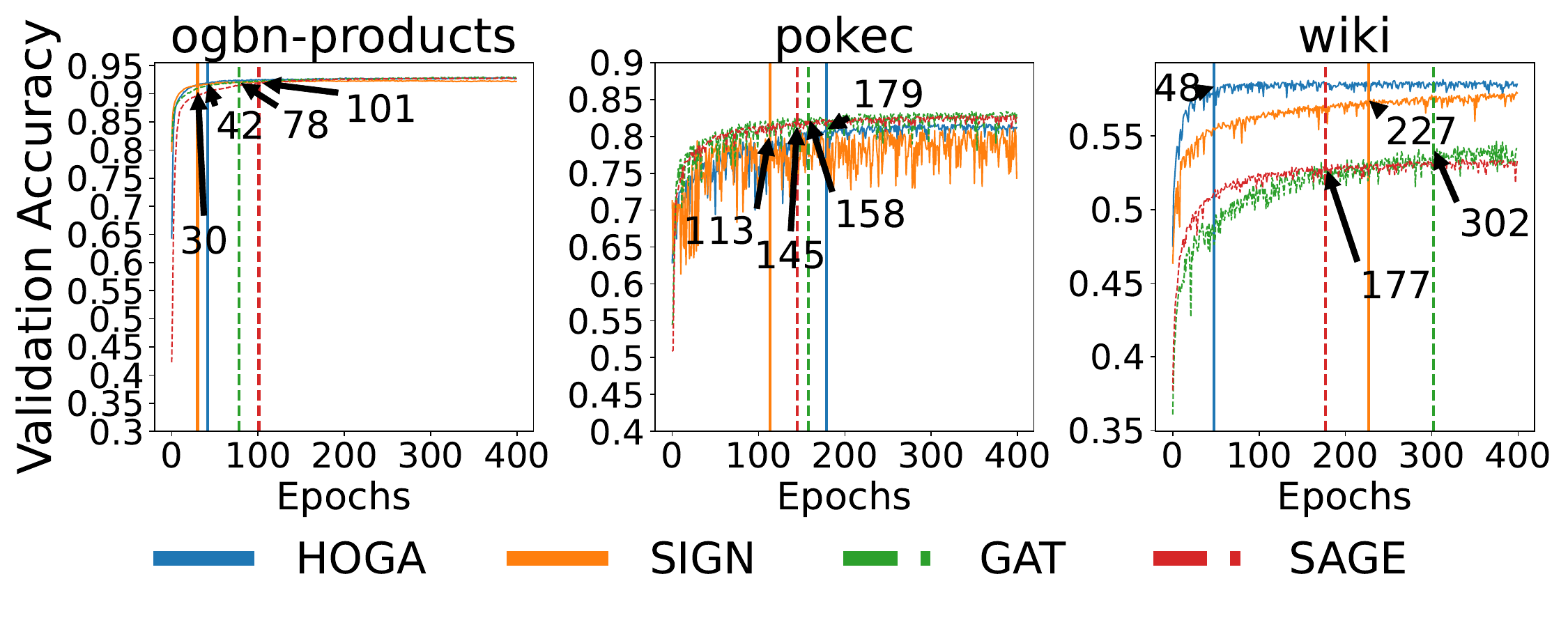}
        \subcaption{Convergence rate comparison under 5 layers or hops.}
        \label{fig:plotC_converge}
    \end{minipage}
    \hfill
    \begin{minipage}[b]{0.49\textwidth}
        \centering
        \includegraphics[width=\textwidth]{ 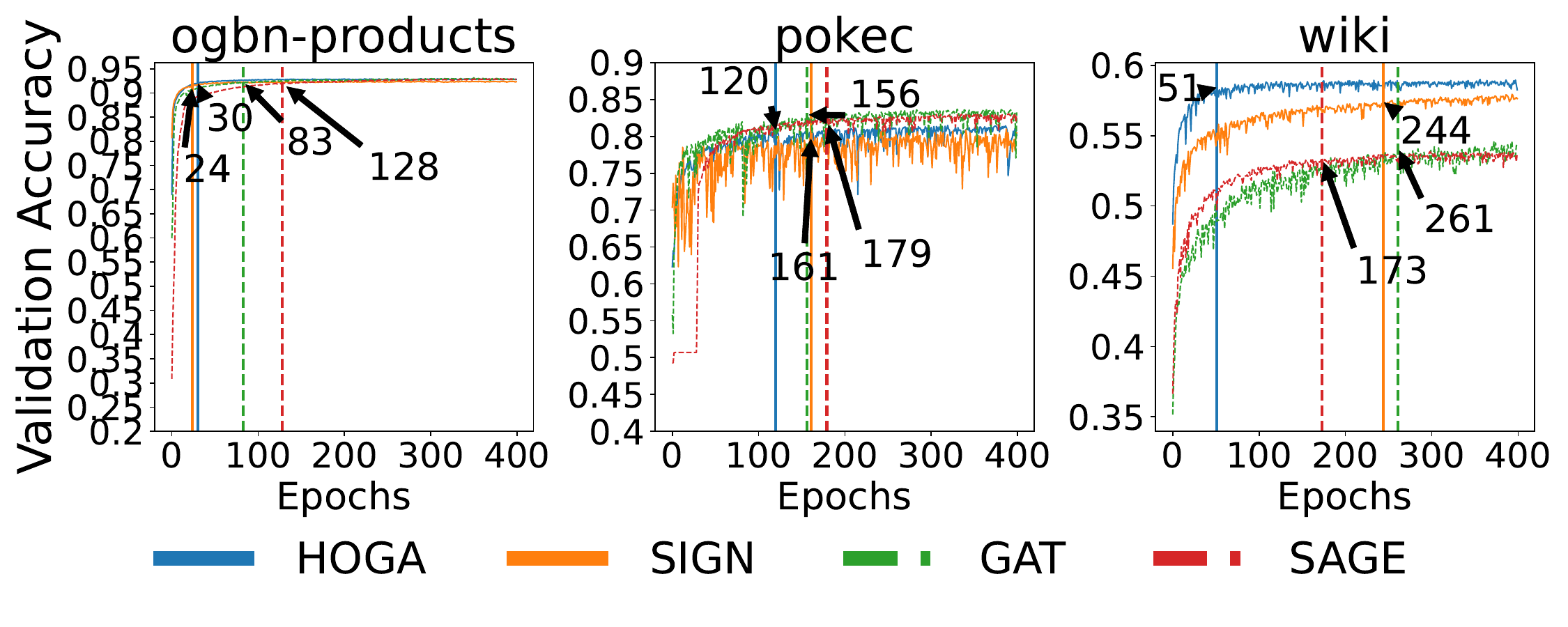}
        \subcaption{Convergence rate comparison under 6 layers or hops.}
        \label{fig:plotD_converge}
    \end{minipage}

    \caption{Convergence rate comparison among \MPGs and \PPGs --- The number in the plot denotes the convergence point where 99\% of peak validation accuracy is reached.}
    \label{fig:convergence_all}
\end{figure*}
We compare the convergence rate of \MPG and \PPG models on three medium-sized datasets under different layers or hops. The results for 2, 3, 5, and 6 layers (hops) are shown in Figure \ref{fig:convergence_all}. In Figure \ref{fig:convergence_all}, we observe that \PPGs consistently converge faster on the \product dataset. On \pokec, the convergence rates of \MPGs are comparable to those of \PPGs. For the \wiki dataset, HOGA achieves the fastest convergence, while GAT converges the slowest, with SIGN performing similarly to GraphSAGE. Overall, \PPGs demonstrate comparable or faster convergence rates than \MPGs, which brings them even more advantages compared to \MPGs when end-to-end training time is considered.

\section{Detailed Experiment Environment}
\label{appendix-software}
We conduct the experiments on a Linux server with two 3.0 GHz Intel Xeon Gold 6248R CPUs (2x24 cores), 380 GB RAM, four RTX A6000 GPUs (each with 48 GB of GPU memory), and two Samsung PM9A3 SSDs (3.5 TB each with 4x PCIe 4.0 support). Regarding software versions, we use PyTorch 2.0.1, DGL 2.1.0, and CUDA 12.1.

\section{Accuracy-Efficiency Trade-off}
\label{appendix-preto}
\begin{figure*}[ht]
    \centering
    \includegraphics[width=0.9\textwidth]{ 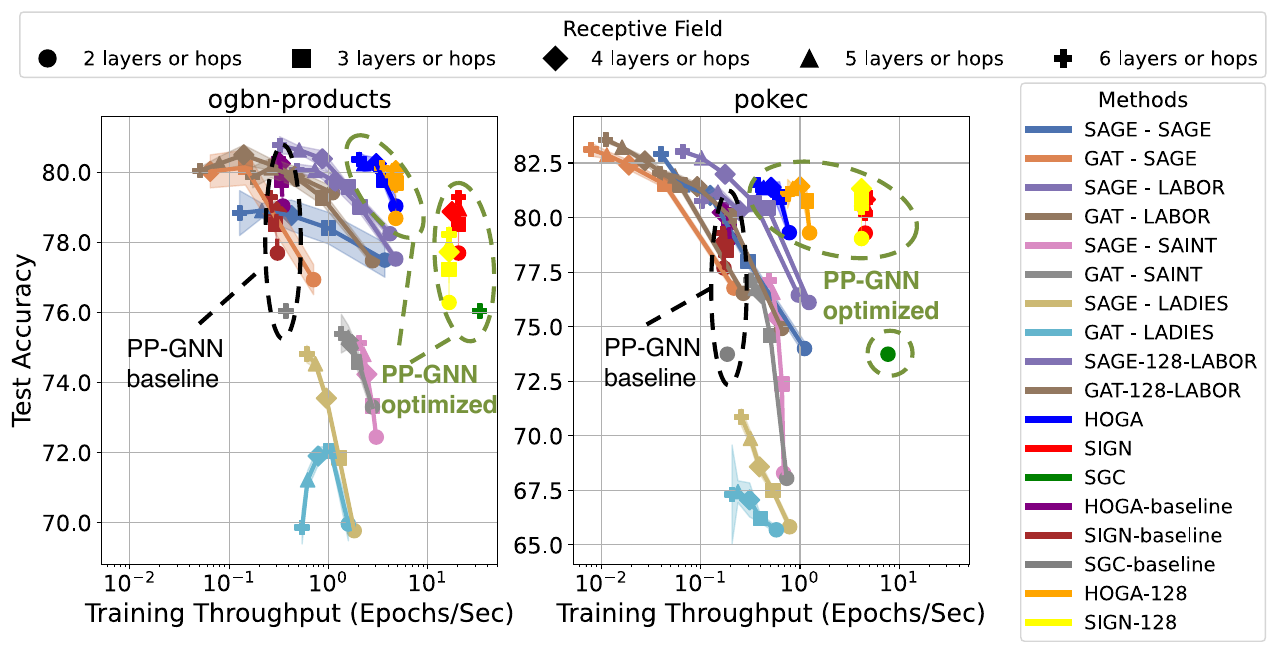}
    \caption{Accuracy efficiency trade-off comparison among \MPGs and \PPGs --- The Y axis represents mean test accuracy obtained from 5 runs, each with 400 epochs. In the legend, for \MPGs, the name to the left of the dash denotes the foundation model (GraphSAGE or GAT), while the name to the right denotes the sampler adopted. The 128 in the labels represents an additional model setting with a hidden dimension of 128 for fair training efficiency comparison.}
    \label{fig:pareto_frontier_products_pokec}
\end{figure*}
The accuracy-efficiency trade-off diagrams on \product and \pokec are shown in Figure \ref{fig:pareto_frontier_products_pokec}. For these experiments, we fine-tune the models as described in Appendix \ref{appendix-parameter}. The accuracy is averaged over 5 runs with 400 epochs each. We observe that \PPGs always lie on the Pareto-Frontier in the diagrams after applying our proposed system-level optimizations, showing significant training efficiency advantage. Regarding accuracy, HOGA and SIGN achieve comparable accuracy as \MPGs with node-wise samplers on these two datasets. 

\section{Chunk Reshuffling}
\label{appendix-chunkreshuffle}
\begin{figure*}[ht]
    \centering
    \begin{minipage}[b]{0.32\textwidth}
        \centering
        \includegraphics[width=\textwidth]{ 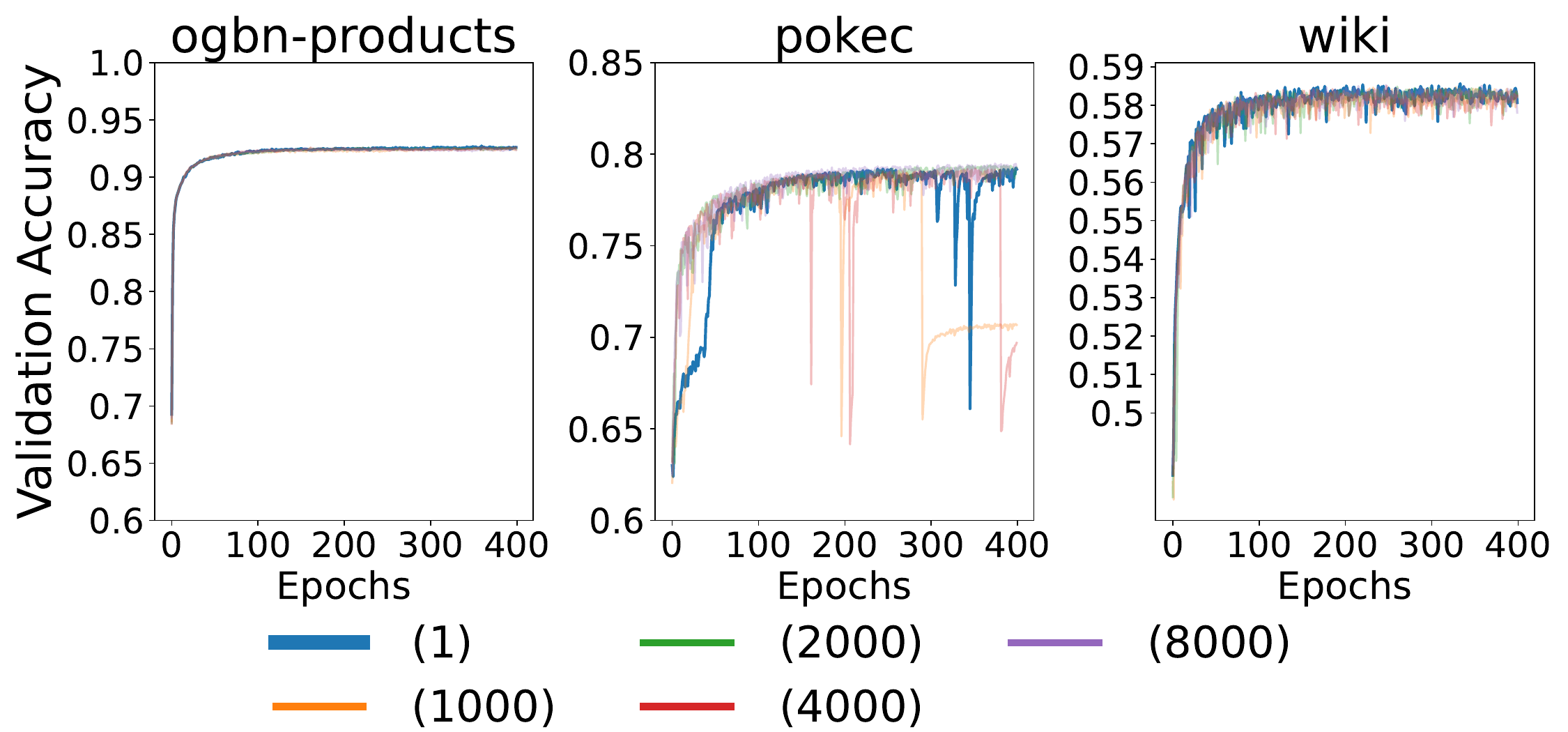}
        \subcaption{Validation accuracy of HOGA with 2 hops}
        \label{fig:plotA}
    \end{minipage}
    \hfill
    \begin{minipage}[b]{0.32\textwidth}
        \centering
        \includegraphics[width=\textwidth]{ 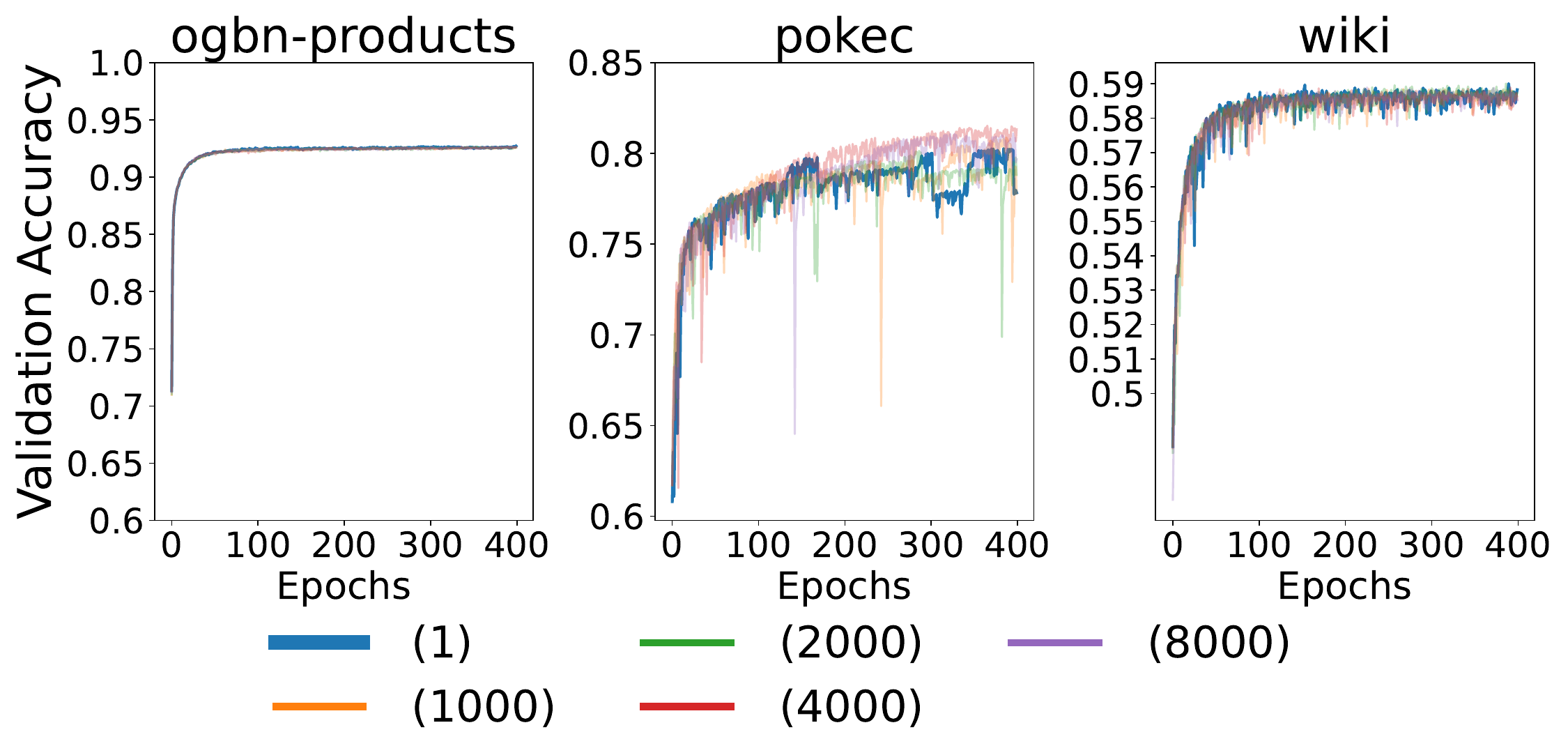}
        \subcaption{Validation accuracy of HOGA with 3 hops}
        \label{fig:plotB}
    \end{minipage}
    \hfill
    \begin{minipage}[b]{0.32\textwidth}
        \centering
        \includegraphics[width=\textwidth]{ 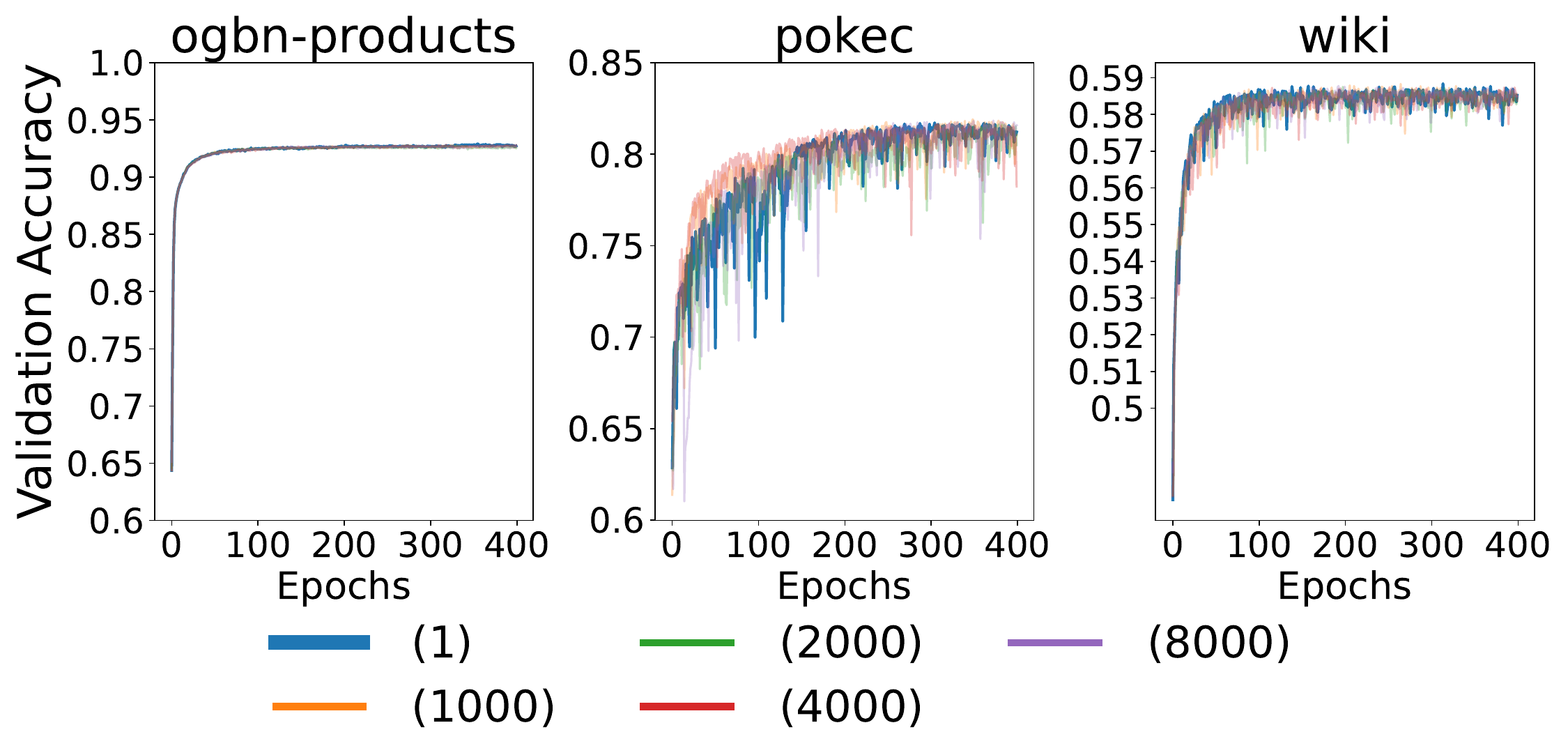}
        \subcaption{Validation accuracy of HOGA with 5 hops}
        \label{fig:plotC}
    \end{minipage}

    \vspace{0.5cm} 

    \begin{minipage}[b]{0.32\textwidth}
        \centering
        \includegraphics[width=\textwidth]{ 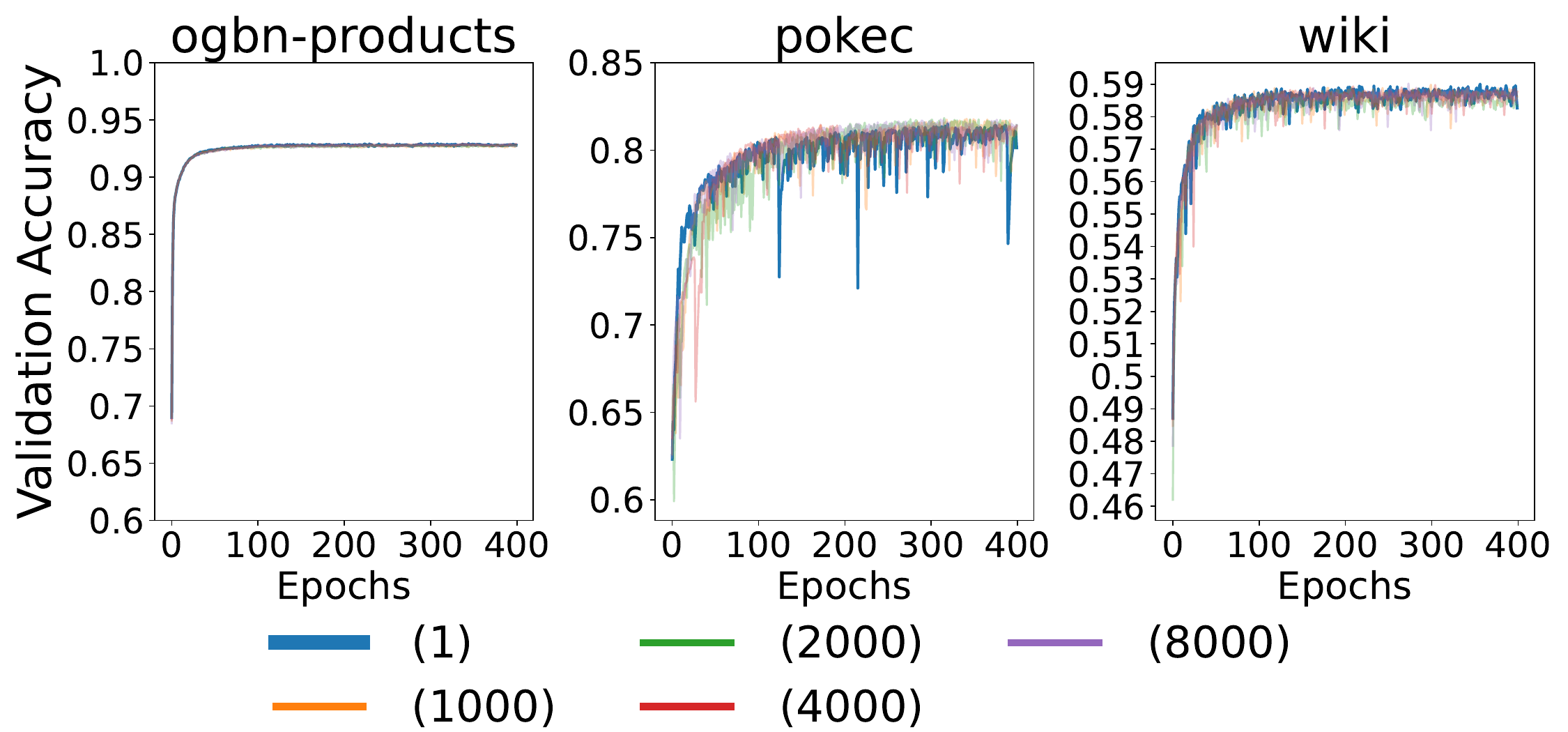}
        \subcaption{Validation accuracy of HOGA with 6 hops}
        \label{fig:plotD}
    \end{minipage}
    \hfill
    \begin{minipage}[b]{0.32\textwidth}
        \centering
        \includegraphics[width=\textwidth]{ 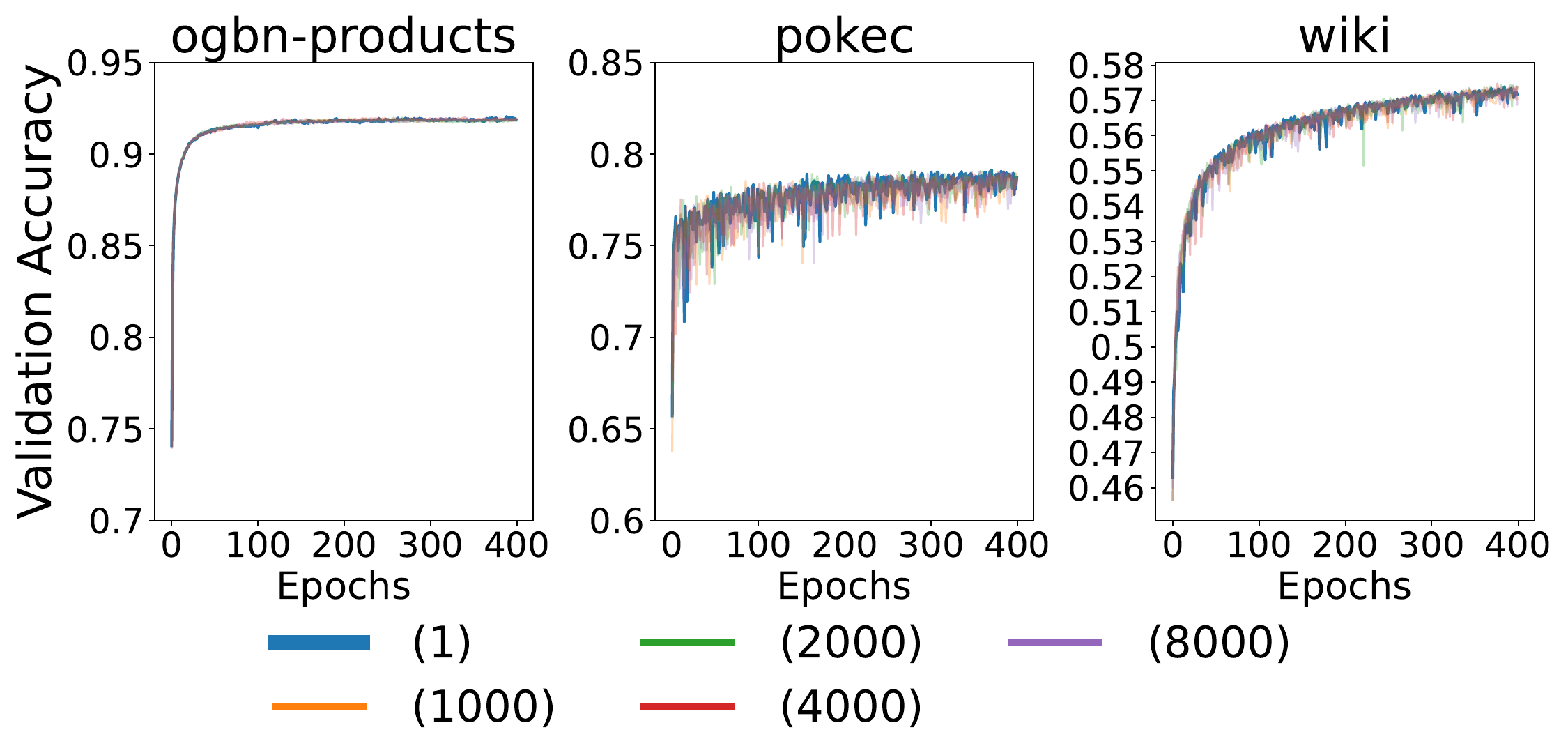}
        \subcaption{Validation accuracy of SIGN with 2 hops}
        \label{fig:plotE}
    \end{minipage}
    \hfill
    \begin{minipage}[b]{0.32\textwidth}
        \centering
        \includegraphics[width=\textwidth]{ 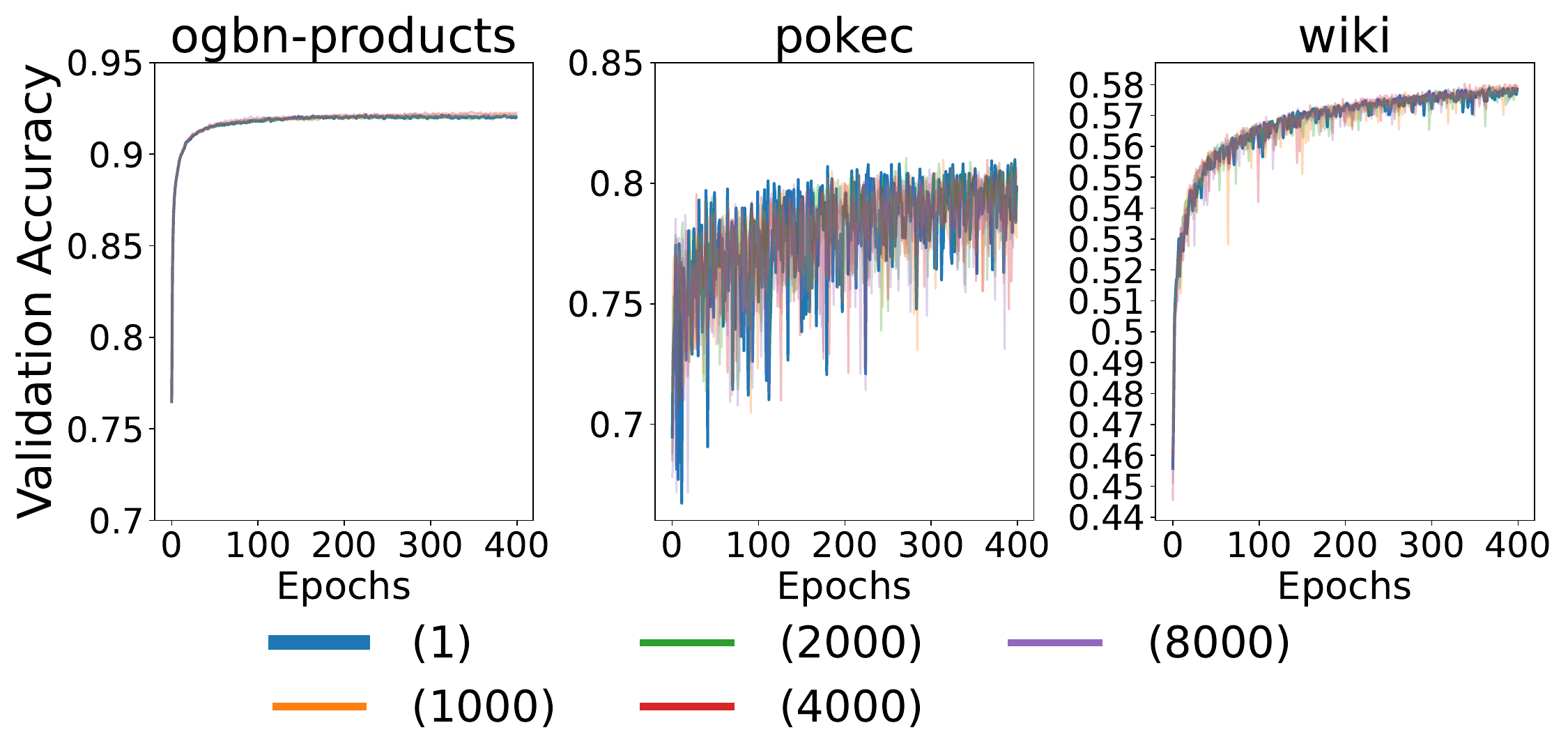}
        \subcaption{Validation accuracy of SIGN with 3 hops}
        \label{fig:plotF}
    \end{minipage}

    \vspace{0.5cm}

    \begin{minipage}[b]{0.32\textwidth}
        \centering
        \includegraphics[width=\textwidth]{ 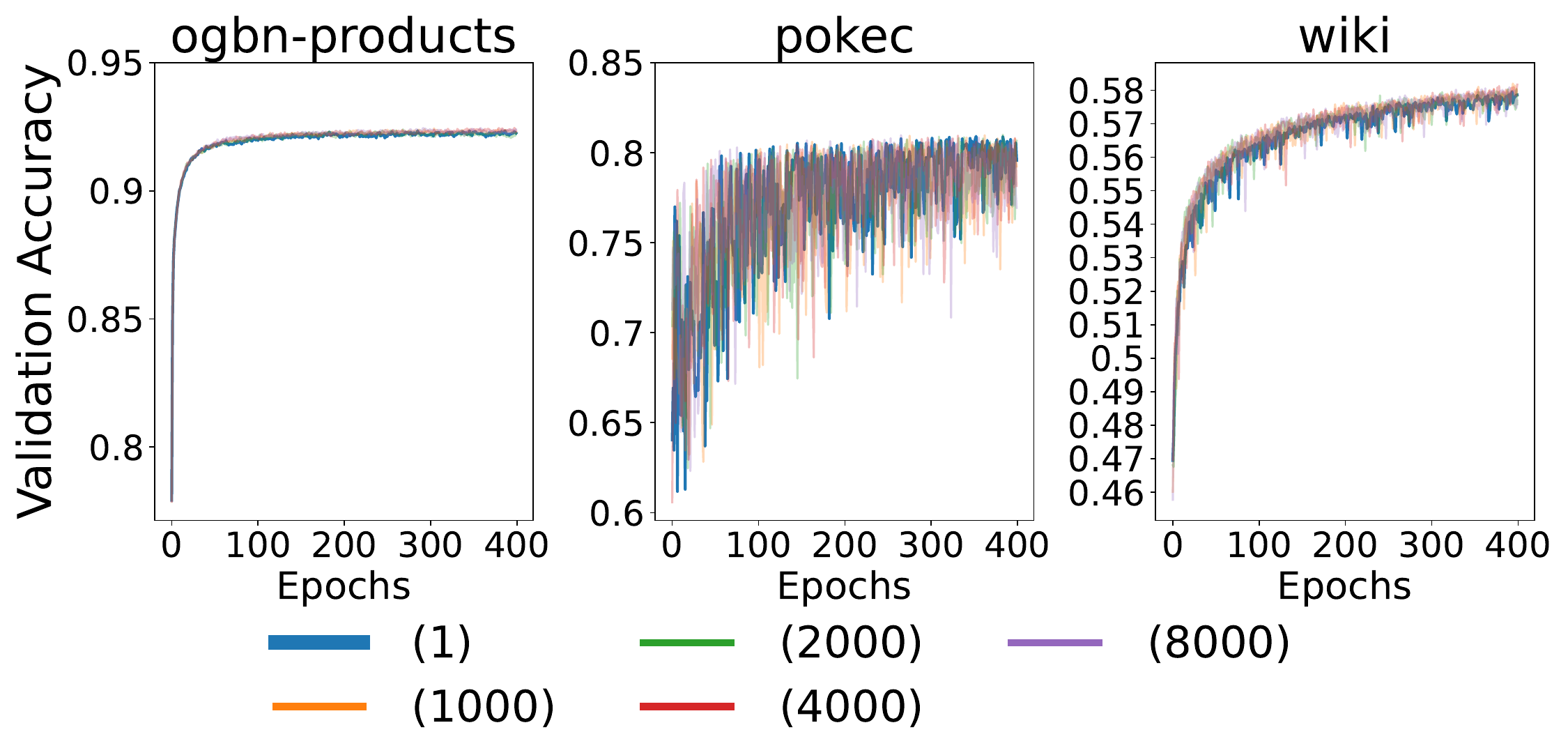}
        \subcaption{Validation accuracy of SIGN with 4 hops}
        \label{fig:plotG}
    \end{minipage}
    \hfill
    \begin{minipage}[b]{0.32\textwidth}
        \centering
        \includegraphics[width=\textwidth]{ 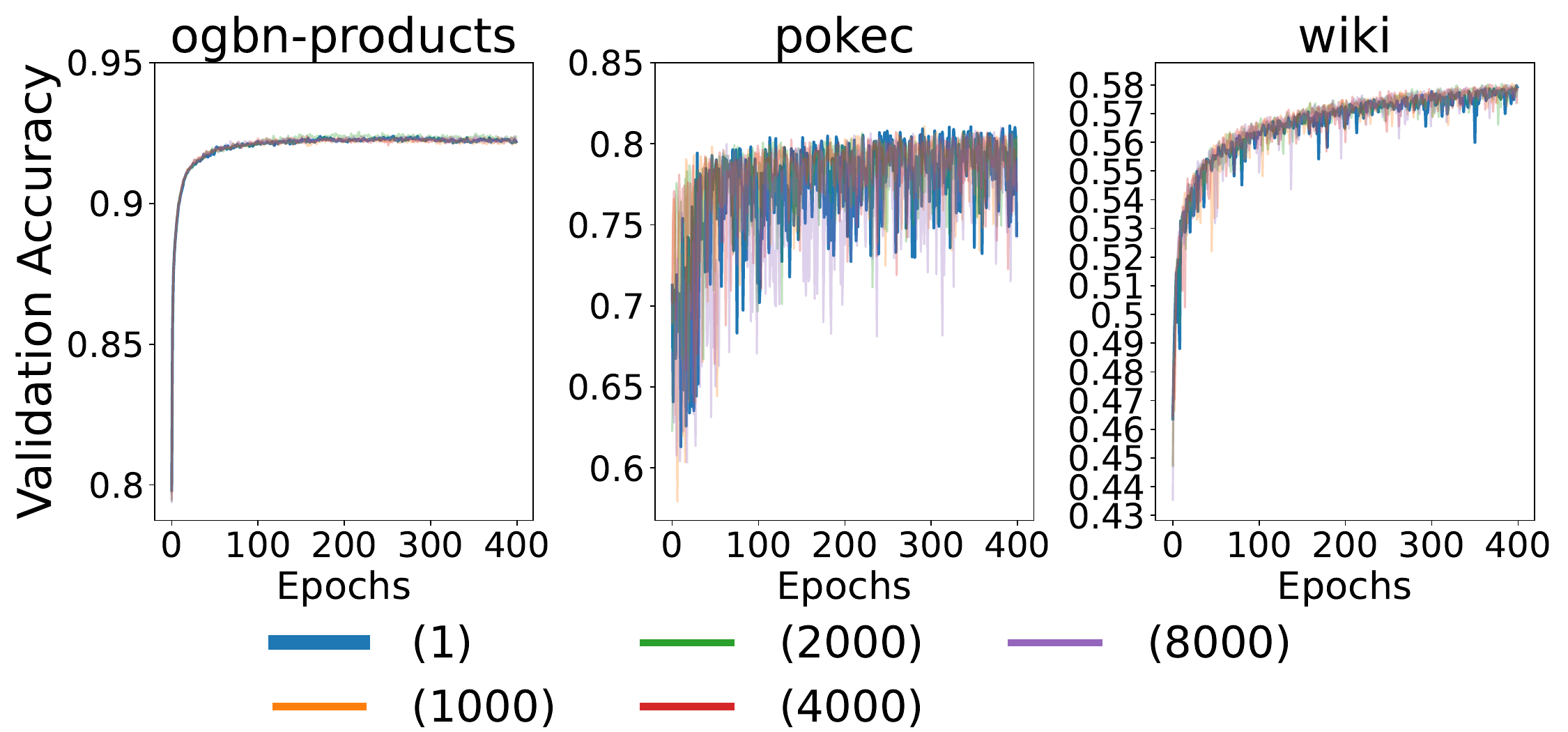}
        \subcaption{Validation accuracy of SIGN with 5 hops}
        \label{fig:plotH}
    \end{minipage}
    \hfill
    \begin{minipage}[b]{0.32\textwidth}
        \centering
        \includegraphics[width=\textwidth]{ 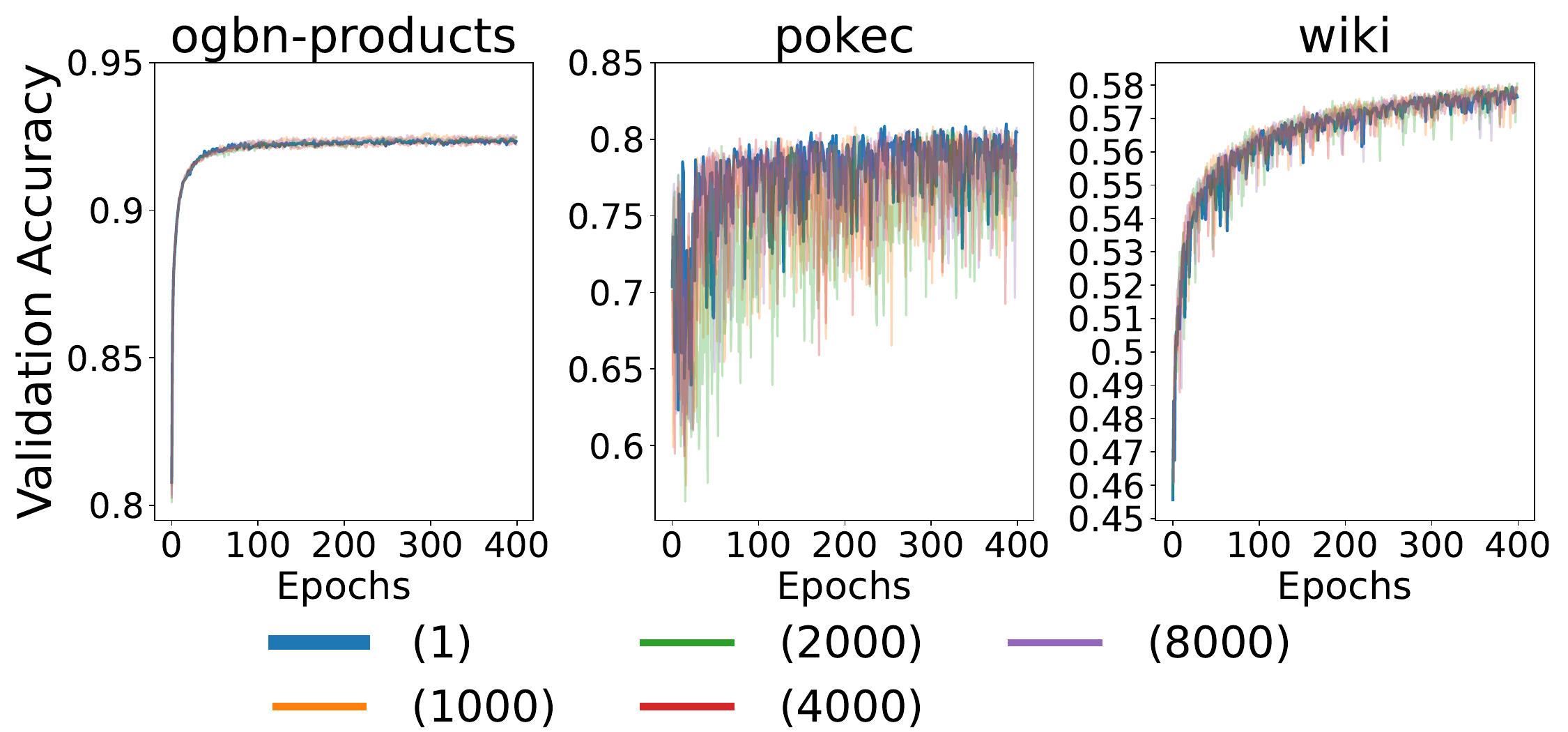}
        \subcaption{Validation accuracy of SIGN with 6 hops}
        \label{fig:plotI}
    \end{minipage}

    \caption{Validation accuracy of HOGA and SIGN under different numbers of hops on three datasets --- The number in the legend denotes the chunk size.}
    \label{fig:chunk_reshuffle_all}
\end{figure*}

\begin{table}[ht]
\centering
\caption{Test accuracy of HOGA and SIGN across different hops and chunk sizes under \pokec.}
\begin{adjustbox}{max width=\columnwidth}
\begin{tabular}{c c c c||c c c c}
\hline
\multicolumn{4}{c||}{\textbf{HOGA}} & \multicolumn{4}{c}{\textbf{SIGN}} \\
\hline
\textbf{Model} & \textbf{Hops} & \textbf{Chunk Size} & \textbf{Acc / \%} & \textbf{Model} & \textbf{Hops} & \textbf{Chunk Size} & \textbf{Acc / \%} \\
\hline
\multirow{5}{*}{HOGA} & \multirow{5}{*}{2} & 1    & 79.32 & \multirow{5}{*}{SIGN} & \multirow{5}{*}{2} & 1    & 79.20 \\
                      &                    & 1000 & 79.21 &                      &                    & 1000 & 79.03 \\
                      &                    & 2000 & 79.43 &                      &                    & 2000 & 79.15 \\
                      &                    & 4000 & 79.25 &                      &                    & 4000 & 79.12 \\
                      &                    & 8000 & 79.58 &                      &                    & 8000 & 79.13 \\
\hline
\multirow{5}{*}{HOGA} & \multirow{5}{*}{3} & 1    & 80.27 & \multirow{5}{*}{SIGN} & \multirow{5}{*}{3} & 1    & 80.96 \\
                      &                    & 1000 & 80.91 &                      &                    & 1000 & 80.95 \\
                      &                    & 2000 & 80.22 &                      &                    & 2000 & 80.92 \\
                      &                    & 4000 & 81.47 &                      &                    & 4000 & 80.94 \\
                      &                    & 8000 & 81.01 &                      &                    & 8000 & 80.55 \\
\hline
\multirow{5}{*}{HOGA} & \multirow{5}{*}{4} & 1    & 81.79 & \multirow{5}{*}{SIGN} & \multirow{5}{*}{4} & 1    & 80.80 \\
                      &                    & 1000 & 81.35 &                      &                    & 1000 & 80.88 \\
                      &                    & 2000 & 81.85 &                      &                    & 2000 & 80.90 \\
                      &                    & 4000 & 81.67 &                      &                    & 4000 & 80.83 \\
                      &                    & 8000 & 81.43 &                      &                    & 8000 & 80.82 \\
\hline
\multirow{5}{*}{HOGA} & \multirow{5}{*}{5} & 1    & 81.65 & \multirow{5}{*}{SIGN} & \multirow{5}{*}{5} & 1    & 81.01 \\
                      &                    & 1000 & 81.90 &                      &                    & 1000 & 80.99 \\
                      &                    & 2000 & 81.64 &                      &                    & 2000 & 80.73 \\
                      &                    & 4000 & 81.68 &                      &                    & 4000 & 80.63 \\
                      &                    & 8000 & 81.91 &                      &                    & 8000 & 80.87 \\
\hline
\multirow{5}{*}{HOGA} & \multirow{5}{*}{6} & 1    & 81.51 & \multirow{5}{*}{SIGN} & \multirow{5}{*}{6} & 1    & 80.90 \\
                      &                    & 1000 & 81.83 &                      &                    & 1000 & 80.67 \\
                      &                    & 2000 & 81.82 &                      &                    & 2000 & 80.72 \\
                      &                    & 4000 & 81.38 &                      &                    & 4000 & 80.54 \\
                      &                    & 8000 & 81.69 &                      &                    & 8000 & 80.64 \\
\hline
\end{tabular}
\end{adjustbox}
\end{table}

\label{tab:chunk_reshuffle_acc_pokec}

We investigate the influence of chunk reshuffling on model convergence rate and accuracy using three medium-sized datasets, with a chunk size chosen from [1, 1000, 2000, 4000, 8000] while all other hyperparameters stay the same as in the accuracy-efficiency tradeoff plots with a single run. The complete results for 2, 3, 5, and 6 hops are shown in Figure \ref{fig:chunk_reshuffle_all}.
From the figure, we observe that the validation accuracy brought by chunk size is negligible on \product and \wiki. On \pokec, the training process is less stable, shown as fluctuations in the training curve, especially for SIGN. However, we find the test accuracy chosen according to the highest validation accuracy is relatively stable, as shown in Table \ref{tab:chunk_reshuffle_acc_pokec}. In the table, a chunk size of 1 equals SGD-RR, and we can see the accuracy degradation brought by the chunk reshuffling training method is less than 0.5\%.

We also examine the effect of chunk reshuffling on a large dataset, \paper, using a chunk size of 8000 under 2, 3, and 4 hops. For HOGA, the test accuracies are 66.09\%, 66.45\%, and 66.75\%, respectively, with a maximum drop of 0.2\% compared to SGD-RR. For SIGN, the test accuracies are 65.55\%, 65.89\%, and 66.12\%, with at most a 0.4\% drop. These results further confirm that SGD-CR has a negligible impact on the accuracy of \PPGs, even on large graphs with over 100 million vertices.

\section{Ablation Study}
\label{appendix-ablationstudy}

As demonstrated in Figure \ref{fig:ablation},
our results highlight the significant benefits of double-buffer prefetching for datasets with larger input feature dimensions, like \wiki, and for simpler \PPG models like SIGN and SGC, where data loading takes a larger portion of training time. When input data is preloaded to GPU memory, our results demonstrate that the double-buffer-based data prefetching scheme offers an average 1.33$\times$ speedup. However, since data loading is no longer the bottleneck due to the high GPU memory bandwidth, further applying chunk reshuffling does not yield additional performance gains.

\section{Preprocessing Overhead}
\label{appendix-preprocessing}
\begin{table*}[ht]
\centering
\caption{Pre-processing Overhead Comparison}
\begin{adjustbox}{max width=0.9\textwidth}
\begin{tabular}{l|c c c c c c}
\hline
 \textbf{Model} & \textbf{\makecell{Number \\ of Hops} } & \textbf{\makecell{Wall Clock \\ Preprocessing \\ Time (sec)}} & \textbf{ \makecell{Epoch \\ Time (sec)}} & \textbf{\makecell{\# Epochs in \\ A Single \\ Training Run}} & \textbf{\makecell{Estimated \\ Training Time of\\ A Single Run (sec)}} & \textbf{ \makecell{Preprocessing Time \\ Compared to A \\ Single Training Run }}\\ \hline
ogbn-products & 6 & 51.8 & 0.49 & 200 & 98 & 53\% \\
pokec & 6 & 27.59 & 2.65 & 400 & 1060 & 3\% \\
wiki & 6 & 122.79 & 2.89 & 400 & 1156 & 11\% \\
igb-medium & 3 & 386.63 & 36.31 & 100 & 3631 & 11\% \\
ogbn-papers100M & 4 & 507.8 & 2.81 & 200 & 562 & 90\% \\
igb-large & 3 & 4521.5 & 539.5 & 30 & 16185 & 28\% \\
\hline
\end{tabular}
\label{table:preprocessing}
\end{adjustbox}
\end{table*}
In Table \ref{table:preprocessing}, we report the wall-clock pre-processing time for the six graph datasets, expressed both in absolute terms and as a proportion of the time for a single training run. The pre-processing step is implemented in PyTorch and leverages a single GPU, except \igbl and \paper, which use CPU only. Consistent with our accuracy evaluations, we utilize only one operator during data pre-processing. Specifically, for \product, \pokec, and \wiki, we extract 6 hops of features; for \paper, 4 hops; and for \igbm and \igbl, 3 hops.

The wall-clock time for a complete training run is estimated by multiplying the per-epoch time of HOGA at the maximum hop number by the total number of epochs (detailed in Table \ref{table:preprocessing}). From Appendix \ref{appendix-converge}, we observe that for \product, 200 epochs suffice for HOGA and SIGN to achieve convergence, while for \pokec and \wiki, 400 epochs are necessary. 

\begin{figure}[ht]
    \centering
    \includegraphics[width=0.95\columnwidth]{ 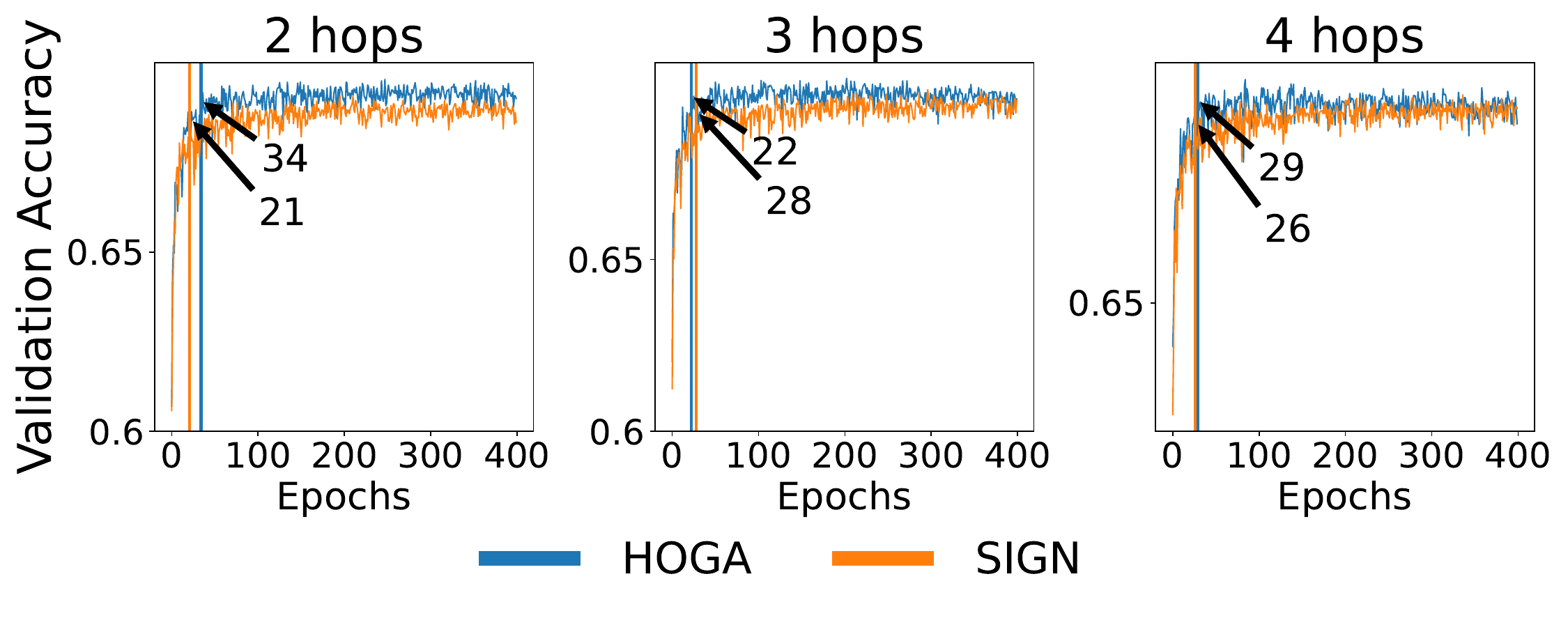} 
    \vspace{-7pt}
    \caption{Convergence rate of HOGA and SIGN on \paper --- The number in the plot denotes the convergence point where 99\% of peak validation accuracy is reached.}
    \label{fig:convergence_paper}
\end{figure}
For larger datasets, we report the test accuracy under 100 epochs, 20 epochs, and 3 epochs for \paper, \igbm, and \igbl, respectively, in Section \ref{subsec:largegraph}, due to the prolonged training time of the \MPG baselines. For \PPGs, we run both HOGA and SIGN for 400 epochs on \paper, and plot their training curves as shown in Figure \ref{fig:convergence_paper}. From the training curves, 200 epochs are enough for HOGA and SIGN to achieve convergence. On \igbm, we further run 100 epochs for HOGA and SIGN, observing a 0.1\% increase in test accuracy, thus selecting 100 epochs for the run time estimation. On \igbl, we run 30 epochs, which yields a test accuracy increase of 0.5\% for HOGA compared to 3 epochs. Therefore, we use 30 epochs as a conservative estimation for the run time. This run time estimation does not account for minor factors, such as data loading time during training, providing an idealized comparison to pre-processing overhead. 

From Table \ref{table:preprocessing}, we observe that the pre-processing overhead is notably lower than that of a single training run, with the exception of the \paper dataset. In \paper, the labeled data is less than 1.4\% of the total nodes, meaning only around one million nodes are used in training. However, the pre-processing step involves all nodes, requiring matrix multiplication across over 111 million nodes, which significantly increases the pre-processing time relative to the epoch time. However, the preprocessing time is still less than the time for a single run of training and can be amortized during the hyperparameter tuning and model adjustment processes, where tens or hundreds of runs are usually required.

\section{Data Placement Study}
\label{appendix-placement}
\begin{figure}[!ht]
    \centering
    \includegraphics[width=\columnwidth]{ 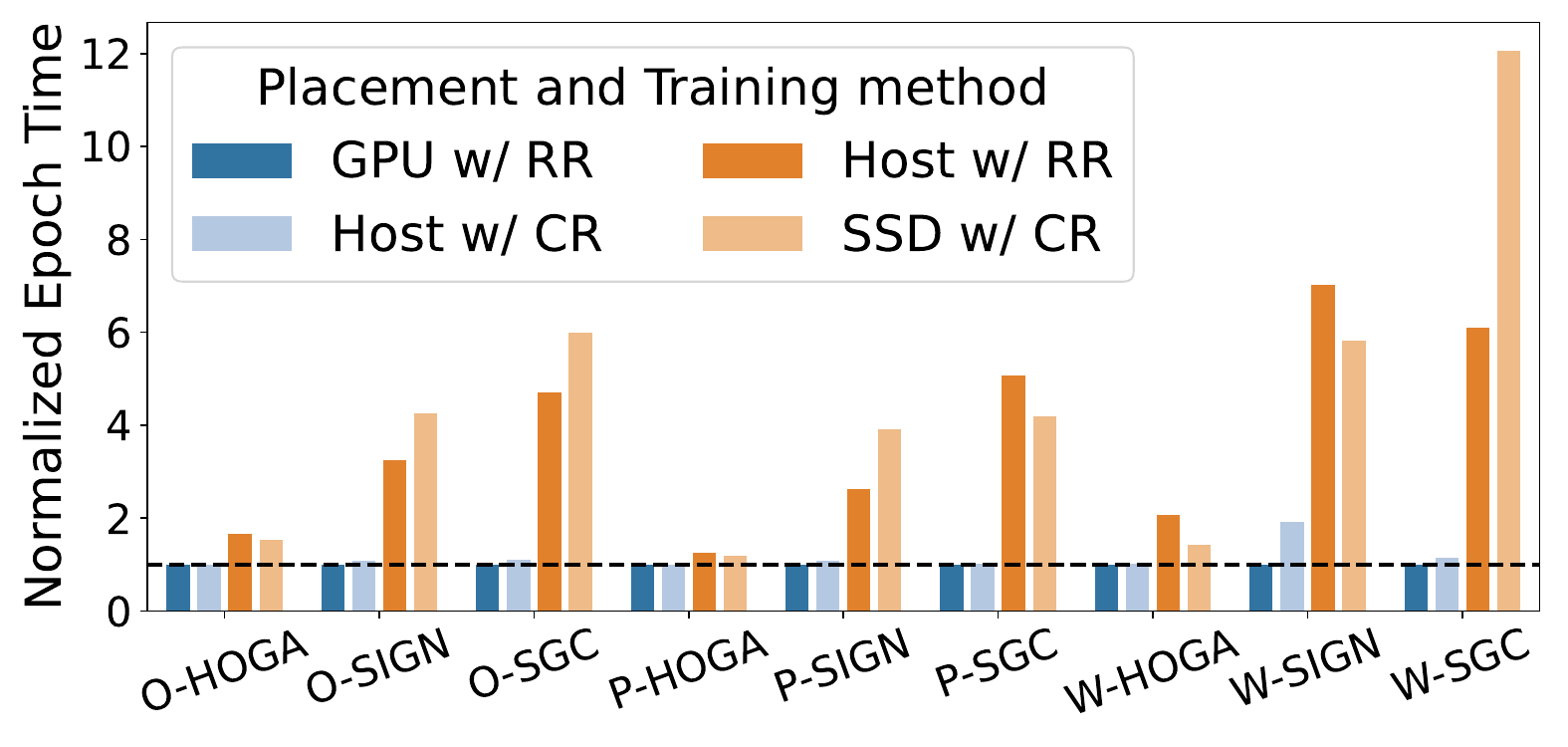}
    \caption{Placement influence on epoch time --- The first item in the legend represents input data location while the second item represents the training method, with RR standing for SGD-RR and CR for chunk reshuffling.}
    \label{fig:placement}
\end{figure}
We evaluate the impact of input data locations on training efficiency to validate our data placement policy. Figure \ref{fig:placement} shows the normalized epoch time for various \PPG models across different datasets, input data locations, and training methods, averaged over 2 to 6 hops and 100 epochs using the geometric mean.

Storing input data in GPU memory maximizes training efficiency due to its high bandwidth. When data is in host memory with chunk reshuffling, the efficiency remains comparable to GPU memory preloading. Using SGD-RR with data in host memory, training time increases moderately for HOGA but significantly for SIGN and SGC compared to chunk reshuffling. This is primarily due to the lighter-weight computation in SIGN and SGC.

When data is read directly from SSD, HOGA’s training time is comparable to or even shorter than when data is read from host memory with SGD-RR. This is due to efficient bulk data transfer enabled by chunk reshuffling and GPU-side double buffering, which largely hides SSD-to-GPU transfer time. However, for SIGN and SGC, data transfer time exceeds GPU computation time, resulting in a notable relative increase in overall training time. On average, direct storage loading achieves 36\% of the training efficiency of GPU memory loading and 41\% of host memory loading with chunk reshuffling, while being 2\% faster than host memory loading with SGD-RR.

\section{Data Transfer Analysis}

We analyze the total data transfer between disk, host, and GPU memory during training. When no caching is applied, \PPG models incur 1–2 orders of magnitude less data transfer compared to \MPG models, highlighting their superior efficiency. This difference arises from the significant node overlap among subgraphs in \MPG training. We profiled the data volume of node features extracted during \MPG subgraph construction as an estimate of data transferred without caching.
 When caching is applied—such as GPU-side caching of data from main memory, or host-side caching of data from disk—the data transfer required by \MPGs can be greatly reduced. However, such caching is less effective for \PPGs, as they do not reuse training data within a single epoch. For \PPGs, the total data transfer volume can be directly estimated from the number of hops used.

The detailed profiling results are as follows:
\begin{itemize}
    \item \textbf{Medium-sized datasets (fitting in GPU memory):} \PPGs require 0.2–15 GB of data transfer, whereas \MPGs require 8$\times$–26$\times$ more.
    \item \textbf{\paper:} \PPGs load less than 3 GB from GPU memory, while \MPGs require 26$\times$–111$\times$ more data transfer from host memory.
    \item \textbf{\igbm:} \PPGs transfer 70–93 GB from host memory, while \MPGs transfer 23$\times$–65$\times$ more.
    \item \textbf{\igbl:} \PPGs transfer 720–960 GB from storage, whereas \MPGs require 16$\times$–55$\times$ more data.
\end{itemize}

These results emphasize the data transfer efficiency of \PPGs. However, data transfer volume does not always directly correspond to training throughput, as models may be either memory-bound or compute-bound. For instance, HOGA and SIGN load the same amount of training data, yet their throughputs differ by more than 10$\times$. When training data is loaded from disk, the throughput advantage of \PPGs more closely aligns with their reduced data transfer volumes compared to \MPGs, suggesting that both GNN families are more likely constrained by storage bandwidth in such scenarios.

\section{Artifact Appendix}

\subsection{Abstract}

This artifact includes the source code for the system-level optimizations introduced in our paper, encompassing efficient batch assembly, double-buffer-based data prefetching, chunk reshuffling, and storage-based training. Additionally, it provides an automated training configuration system. 

Execution requires a machine with multiple NVIDIA GPUs, NVIDIA GPU Direct Storage (GDS), and an SSD. The artifact includes installation scripts for all dependencies. Due to the computational and storage demands of large graph benchmarks, we provide reproduction instructions for experiments on the \texttt{ogbn-products} dataset. Experiments on other datasets follow similar procedures.

\subsection{Artifact Check-List (Meta-Information)}

{\small
\begin{itemize}
  \item \textbf{Algorithm:} Graph Neural Networks (GNNs)
  \item \textbf{Dataset:} \texttt{ogbn-products}
  \item \textbf{Hardware:} x86 CPU, multiple NVIDIA GPUs, SSD
  \item \textbf{Execution:} Bash scripts for data preprocessing and training
  \item \textbf{Metrics:} Training throughput, accuracy
  \item \textbf{Output:} Standard output (stdout), log files
  \item \textbf{Experiments:} Single-GPU and multi-GPU training, automated training configuration
  \item \textbf{Disk Space Requirement:} ~10 GB
  \item \textbf{Workflow Preparation Time:} ~30 minutes
  \item \textbf{Experiment Completion Time:} ~1 hour
  \item \textbf{Publicly Available:} Yes
  \item \textbf{Code License:} MIT License
  \item \textbf{Frameworks Used:} PyTorch, DGL, PyG
  \item \textbf{Archived (DOI):} TBD
\end{itemize}}

\subsection{Description}

\subsubsection{Delivery Method}
The artifact is available as a GitHub repository:
\begin{itemize}
    \item Repository: \sloppy\url{<https://github.com/cornell-zhang/preprop-gnn>}
\end{itemize}

\subsubsection{Hardware Dependencies}
\begin{itemize}
    \item x86 CPU
    \item Multiple NVIDIA GPUs
    \item SSD
\end{itemize}

\subsubsection{Software Dependencies}
\begin{itemize}
    \item NVIDIA GDS (1.6.0 or higher)
    \item Python 3.9
    \item CUDA 11.8
    \item PyTorch 2.2.1
    \item DGL 2.1.0
    \item PyG 2.5.2
    \item OGB 1.3.6
    \item IGB 0.1.0
\end{itemize}

\subsubsection{Datasets}
\begin{itemize}
    \item \texttt{ogbn-products} (\texttt{pokec}, \texttt{wiki}, \texttt{ogbn-papers100M}, \texttt{IGB-medium}, and \texttt{IGB-large} supported)
\end{itemize}

\subsection{Installation}
\begin{enumerate}
    \item Create a conda environment and install dependencies using the provided script.
    \item Install \texttt{igb} from its official GitHub repository.
    \item Install two custom operators: \texttt{async\_fetch} and \texttt{gds\_read}.
    \item Detailed instructions are provided in the \texttt{README.md}.
\end{enumerate}

\subsection{Experiment Workflow}

The workflow consists of four main parts:
\begin{enumerate}
    \item \textbf{Preprocessing:} Convert the dataset into a format suitable for PP-GNN training.
    \item \textbf{Single-GPU Experiments:} Compare vanilla PP-GNN training with our optimized pipeline, evaluating different data placements:
    \begin{itemize}
        \item In GPU memory
        \item In host memory using SGD-RR or SGD-CR
        \item In storage
    \end{itemize}
    \item \textbf{Multi-GPU Experiments:} Evaluate training with data in GPU and host memory using SGD-RR and SGD-CR. Note that multi-GPU training does not support SGD.
    \item \textbf{Automated Training Configuration Experiments:} Test our automated system for optimizing training configurations.
\end{enumerate}

\subsection{Evaluation and Expected Results}

\begin{itemize}
    \item Accuracy results are stored in \texttt{./result}. For HOAG with 3 hops under 400 epochs with the default settings, the test accuracy should be around 79.7\%.
    \item Training throughput results are stored in \texttt{./result/timing}.
    \item Expected training throughput ranking (single GPU):
    \begin{center}
        \texttt{GPU preloading} $\approx$ \texttt{Host memory with SGD-CR} $>$ \texttt{Host memory with SGD-RR} $\approx$ \texttt{Storage}
    \end{center}
    \item Multi-GPU scalability depends on the hardware configuration.
\end{itemize}

\subsection{Experiment Customization}

\begin{itemize}
    \item Modify \texttt{model\_cfg.json} to explore different models and hyperparameter settings. For instance, change method to SIGN or SGC to explore these two models, change training\_hops to other numbers to exploring using different hops.
    \item Update \texttt{evaluation.sh} to change GPU IDs and \texttt{GPUcap} parameters to use different number of GPUs.
\end{itemize}
\end{document}